\documentclass[nohyperref]{article}

\usepackage{microtype}
\usepackage{graphicx}
\usepackage{subfigure}
\usepackage{booktabs} 

\usepackage[hidelinks]{hyperref}

\usepackage[accepted]{icml2023}

\usepackage{amsmath}
\usepackage{amssymb}
\usepackage{mathtools}
\usepackage{amsthm}
\usepackage{bbm}

\usepackage{color,soul}
\usepackage{colortbl}
\usepackage{balance}
\usepackage{multirow}
\usepackage{multicol}
\newcommand{\specialcell}[2][c]{%
  \begin{tabular}[#1]{@{}c@{}}#2\end{tabular}}

\usepackage{enumitem}

\usepackage[capitalize,noabbrev]{cleveref}

\theoremstyle{plain}

\theoremstyle{definition}

\theoremstyle{remark}

\usepackage[textsize=tiny]{todonotes}

\graphicspath{{figures/}}

\newcommand{\figGap}[0]{\vspace{-1.1\baselineskip}}

\definecolor{lightgreen}{rgb}{0.8,1,0.8}
\definecolor{lightyellow}{rgb}{1,0.95,0.8}
\newcommand{\tbcolorg}{\cellcolor{lightgreen}}

\icmltitlerunning{On the Effectiveness of Offline RL for Dialogue Response Generation}

\begin{document}

\twocolumn[
\icmltitle{On the Effectiveness of Offline RL for Dialogue Response Generation}




\begin{icmlauthorlist}
\icmlauthor{Paloma Sodhi}{comp}
\icmlauthor{Felix Wu}{comp}
\icmlauthor{Ethan R. Elenberg}{comp}
\icmlauthor{Kilian Q. Weinberger}{comp,sch}
\icmlauthor{Ryan McDonald}{comp}
\end{icmlauthorlist}

\icmlaffiliation{comp}{ASAPP, New York, United States}
\icmlaffiliation{sch}{Cornell University, New York, United States}

\icmlcorrespondingauthor{Paloma Sodhi}{psodhi@asapp.com}

\icmlkeywords{text generation, reinforcement learning, dialogue systems}

\vskip 0.3in
]



\printAffiliationsAndNotice{Code available at {\url{https://github.com/asappresearch/dialogue-offline-rl}} \\ \\}  

\begin{abstract}
A common training technique for language models is teacher forcing (TF). TF  attempts to match human language exactly, even though identical meanings can be expressed in different ways. This motivates use of sequence-level objectives for dialogue response generation. In this paper, we study the efficacy of various offline reinforcement learning (RL) methods to maximize such objectives. We present a comprehensive evaluation across multiple datasets, models, and metrics. Offline RL shows a clear performance improvement over teacher forcing while not inducing training instability or sacrificing practical training budgets.

\vspace{-1.1em}

\end{abstract}

\section{Introduction}
\label{introduction}


Dialogue response generation is an important task in natural language processing with numerous applications such as virtual personal assistants and call center agent tools~\cite{zhou2017end, swanson2019building, jaques2020human, ramakrishnan2022long, ouyang2022training}. Historically, text generation models have typically been trained with teacher forcing (TF) \cite{williams1989learning}, which involves predicting the next token in a sequence to exactly match the human utterance in a ground truth dataset. However, this can be a needlessly challenging objective, as a human may choose to say the same thing in multiple different ways. Consider a dialogue system that provides suggestions to an agent during a conversation with a customer. These suggestions need only be \emph{close enough} for an agent to select it. This suggests a different objective, one that is defined on the entire sentence rather than individual tokens.



One way to design such a loss would be to incorporate human-in-the-loop feedback if a model generated utterance matches the meaning of the ground truth sentence. 
However, this can be expensive to collect. Instead, model-based metrics to measure utterance similarity, such as BERTScore~\cite{zhang2019bertscore} and BLEURT~\cite{sellam2020bleurt}, provide a cheaper alternative. These are automated metrics that capture semantic similarity between sentences and tend to have a high correlation with human judgment~\cite{zhang2019bertscore,sellam2020bleurt}. Given a choice of such metrics, what learning framework would allow us to maximize them for dialogue text generation?



Recent works have explored online RL methods for text generation \cite{ranzato2015sequence, li2016deep, ouyang2022training, ramamurthy2022reinforcement}, leading to some exciting successes~\cite{ouyang2022training}. 
However, \textit{offline} RL has received relatively less attention~\cite{jaques2020human, pang2020text}.
We argue that offline RL~\cite{levine2020offline} does provide a framework that meets all aforementioned desiderata. Unlike teacher forcing, it can handle losses on the entire sequence as a reward function and  unlike online RL, it can leverage existing data without having to explore, matching similar training times as teacher forcing. 

In this paper, we present a comprehensive evaluation of offline RL methods for dialogue text generation and investigate best practices. 
We explore three complementary approaches. 
The first, TF Top, is to fine-tune a model on utterances that accrue high returns.
The second, Decision Transformers (DT)~\cite{chen2021decision}, is to train a conditional model that conditions on returns, and at inference time condition on a high return.
The third, ILQL \cite{kostrikov2021offline, snell2022ilql}, is an off-policy Q-learning approach that uses dynamic programming to train a  critic.
All three of these approaches are complementary and have been shown to be competitive outside of dialogue settings, making them great candidates to evaluate the efficacy of offline RL for dialogue text generation.



To summarize our contributions, we formalize three state-of-the-art offline RL approaches for the task of dialogue text generation. We evaluate them across multiple data sets, models, and metrics and provide a thorough ablation analysis of these approaches. We find that offline RL methods show a clear performance improvement over teacher forcing and achieve a trade-off where they generate text close enough in meaning to human. Through different experiments, we demonstrate that the offline RL framework provides an ideal fit for the task of dialogue generation, and should be considered seriously by the community.

\vspace{-0.25em}

\section{Related Work}
\label{related_work}

\begin{figure*}[!t]
\centering
\includegraphics[width=0.95\textwidth, trim = 0cm 0.1cm 0cm 0cm]{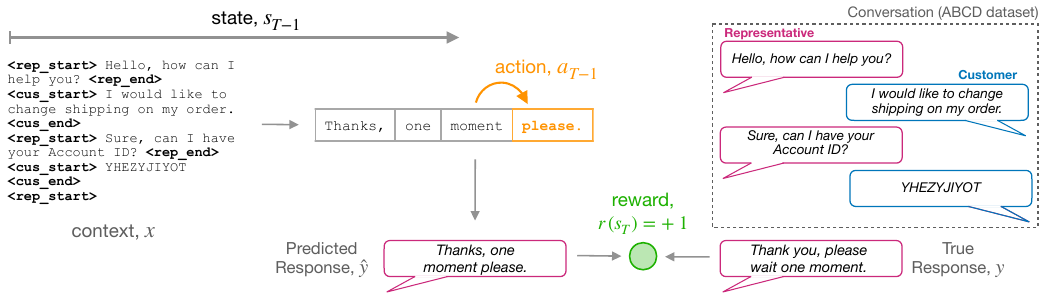}
\caption{\textbf{Dialogue generation as a Markov Decision Process (MDP)} Given a dataset of context-response pairs, each pair is mapped to an MDP episode. States are context $x_i$ and partially generated response up to time $t$, actions are next tokens from vocabulary $V$, and rewards are computed by comparing generated response $\hat{y}$ against target response once $t$ exceeds $T$ or end of sentence token is generated.}
\label{fig:prob_form}
\figGap
\end{figure*}


\paragraph{RL for NLP.}  Prior work has used RL techniques to improve models in a variety of NLP applications~\cite{ranzato2015sequence, pang2020text, yang2020improving, lu2022quark, snell2022ilql, ramamurthy2022reinforcement} such as machine translation~\cite{yonghui2016bridging, wu2018adversarial, kiegeland2021revisiting}, summarization~\cite{paulus2017deep, pasunuru2018multi, stiennon2020learning}, question answering~\cite{furman2022qait}, visual reasoning ~\cite{wu2022lilgym} and instruction following ~\cite{misra2015reinforcement, ouyang2022training}. Techniques adopted range from online RL methods like REINFORCE~\cite{williams1992simple} and PPO~\cite{schulman2017proximal} to offline RL approaches like conservative Q-learning (CQL)~\cite{kumar2020conservative} and decision transformers~\cite{chen2021decision}.

\textbf{RL for Dialogue Generation.} 
Dialogue generation can be challenging as generated sequences can be long and in each turn there can be multiple acceptable responses. \citet{li2016deep} use REINFORCE to optimize a set of rewards that capture informativity, coherence, and ease of answering. \citet{zhou2017end} use a mixture of on and off-policy policy gradient to optimize a reward that captures both utterance-level and dialog-level rewards. \citet{jaques2019wayoff,jaques2020human} use offline RL to optimize a learned reward function from human responses. \citet{ouyang2022training} use PPO to optimize a learned reward model from human ranking. Our goal is to generate dialogue responses that are semantically close to ground truth utterances without having to design explicit rewards that capture dialogue success~\citet{liu2016not}. This is complementary to approaches that look at optimizing dialogue-level metrics like key values for slots \cite{lee2021dialogue, tian2021amendable, bang2023task}.

\textbf{Offline RL for NLP.} Offline RL removes the need for interaction during train time operating only on static datasets of prior human interaction, which leads to improved training stability. \citet{pang2020text} use importance weighted REINFORCE, which only trains a policy without a critic to control for variance. \citet{verma2022chai} use CQL but operate on entire utterances and not per token thus reasoning over shorter sequences. \citet{jaques2019wayoff, jaques2020human} operate at per-token level using off-policy Q-learning, but require generation at RL training time that can be expensive. \citet{snell2022ilql} propose ILQL, a variant of CQL with implicit dataset support constraints, that requires no such generation at train time. \citet{lu2022quark} propose Quark, that uses Decision Transformers by quantizing rewards. While both papers explore metrics like toxicity and sentiment, they don't optimize for similarity to human utterances in dialogue settings that we examine in this paper.

\section{Problem Formulation}
\label{prob_form}

\subsection{Dialogue Response Generation as an MDP}
We look at the problem of dialogue text generation, \textit{i.e.}, generating response utterances in a dialogue setting. We begin with a supervised dataset of context response pairs $\{ x^i, y^i \}_{i=1}^{N}$, where context $x$ is the conversation history, and response $y = \{y_1, \dots, y_T \}$ is a target sequence of tokens. We map each data point $(x, y)$ to an episode of a Markov Decision Process (MDP), which we define below (Fig. \ref{fig:prob_form}):
\begin{itemize}[nosep]
\item \textbf{States, $s_t \in \mathcal{S}$} is the context $x$ and the partially generated sequence of tokens up to and including time step $t$, $\hat{y}_{<t}:=\{\hat{y}_1,\dots, \hat{y}_t\}$. 
\item \textbf{Actions, $a_t\in \mathcal{A}$} are the set of next tokens $\hat{y}_{t+1}$ available from the vocabulary $V$
\item \textbf{Transition function, $\mathcal{T}(s_{t+1} | s_t, a_t)$} is deterministic since every state-action pair $(\hat{y}_{<t}, \hat{y}_{t+1})$ leads to a unique state $\hat{y}_{<t+1}$ for the next step.
\item \textbf{Rewards, $r_t: \mathcal{S}\times\mathcal{A}\rightarrow [0, 1]$} is a terminal reward that computes similarity generated response $\hat{y}$ and 
target response $y$
\item \textbf{Horizon, $T$} is time horizon. Each episode ends when the current time step $t$ exceeds $T$ or an end-of-sentence (EOS) token is generated.
\end{itemize}
The goal is to learn a policy $\pi: s_t \rightarrow a_t$ maximizing \textit{return}, \emph{i.e.} the cumulative reward over an episode $\mathbb{E}_{\pi} \sum_{t=0}^T \gamma^t r_t$. We assume undiscounted cumulative rewards, \emph{i.e.} $\gamma=1$.

\subsection{Rewards for Dialogue Response Generation}
We define the reward to be a similarity metric between the generated text $\hat{y}$ and the speaker's ground truth utterance $y$. Such a metric should capture both what the speaker is trying to communicate and the relevance to the conversation. 

One option is to collect human-in-the-loop annotations, \emph{i.e.} what the speaker would likely prefer to say. However, this requires costly human supervision. Automated metrics, such as BERTScore \cite{zhang2019bertscore}, BLEURT \cite{sellam2020bleurt}, offer a promising alternative. They are able to capture models of human preference and are cheap to evaluate. 

We use a terminal reward since the similarity can only be evaluated at the end of the utterance, and since the same content can be expressed in different styles, \emph{e.g.} \emph{The flight from New York to Boston has been confirmed.} vs. \emph{Your JFK to BOS flight has been booked.}

\subsection{Why Offline Reinforcement Learning?}
\label{sec:why_offline_rl}

In online reinforcement learning, an agent learns by interacting with an environment in real-time. This presents an explore-exploit trade-off, where the agent must balance the need to try out new actions to learn about the environment with the need to exploit its current knowledge to maximize reward. This can be particularly challenging in text generation, as action space (\emph{i.e.} vocabulary size) is often large, \emph{e.g.} of the order of 50,000 words for GPT-based models~\cite{radfordlanguage}. Another problem is that the reward landscape is sparse, hence policies during training can get stuck in local minima where reward is persistently zero. 

For text generation, we argue that an offline setting is reasonable. There exists good generation policies, e.g. policies from teacher forcing, that can generate a set of responses such that one of them is close enough to the human response. Also, once a token is generated, we \emph{deterministically} transition to next state with the additional token appended to the prefix, \emph{i.e.} no interaction is needed to learn the environment.
 
Offline RL provides a learning paradigm that combines both supervised learning’s ability to leverage existing data with the general utility optimization power of online reinforcement learning methods. We collect an offline dataset of state transitions $\mathcal{D} = \{(s_t^i, a_t^i, r_t^i, s_{t+1}^i)\}_{i=1}^{N}$\footnote{We suppress superscripts when considering a single transition.} using a behavior policy $\pi_{\beta}$, typically a policy trained via supervised learning. The goal is to learn a policy $\pi$ that maximizes performance on the dataset while staying close to the behavior policy:
    \begin{align}
        \max_{\pi} J_{\mathcal{D}}(\pi) - \alpha D(\pi, \pi_{\beta}) \, ,
    \end{align}
where $J_{\mathcal{D}}(\cdot)$ is performance on dataset $\mathcal{D}$ and $D(\cdot, \pi_{\beta})$ is distributional regularization against behavior policy $\pi_{\beta}$.

\section{Approach}
\label{approach}

In this section we introduce and compare three recent approaches to offline RL. For all methods, we begin with a pre-trained language model $\pi_\beta$ trained via teacher-forcing, and use this to generate the offline dataset $\mathcal{D}$. 

\subsection{Fine Tune on Top Returns}
The simplest approach is to fine-tune a model on ``top'' demonstrations, \textit{i.e.} teacher forcing on top returns (TF-Top) 

We define a subset of the dataset $\mathcal{D}_{\mathrm{top}}$ that has high returns above a specified threshold, where return is the cumulative reward until the end of the episode, $\hat{Q}(s_t, a_t) = \sum_{t}^T r_t$. The gradient update is simply the log-likelihood gradient on the data subset $\mathcal{D}_{\mathrm{top}}$,
\begin{equation}
\begin{aligned}
&\underset{s_t, a_t \sim \mathcal{D}_{\mathrm{top}}}{\mathbb{E}} \left[ \nabla_{\theta} \log \pi_\theta (a_t | s_t) \right] \, , \\
 \text{where  }\; & \mathcal{D}_{\mathrm{top}} = \{ (s_t, a_t) \in \mathcal{D} \; | \; \hat{Q}(s_t, a_t) \geq 1 - \delta \} \, .
\end{aligned}
\end{equation}
Here, $\delta$ is a specified threshold defining a ``good enough'' return. $\delta$ can be computed by taking the top percentile of all returns $\hat{Q}(s_t, a_t)$ \footnote{Note that $\hat{Q}(s_t, a_t) \leq 1$.}. Since we use a terminal undiscounted reward, the return for any token along the sequence is the same as the final reward received at the end of the sequence. Additionally, if the reward is binary $\{0, 1\}$, $\mathcal{D}_{\mathrm{top}}$ selects sequences corresponding to the reward $1$.

However, one artifact of this approach is that it only increases likelihood of ``good'' tokens, but doesn't necessarily decrease the likelihood of ``bad'' tokens. This is because we \emph{discard} trajectories with low return that were likely under the original TF policy $\pi_{\beta}$, rather than using them to update the model's parameters in the opposite direction. 


\subsection{Decision Transformers: Condition on Return}

Decision Transformer (DT) \cite{chen2021decision} is an approach that reduces offline reinforcement learning to supervised learning. The core idea of DT is to learn the return-conditional distribution of actions in each state, and then define a policy by sampling from the distribution of actions that receive high returns.

Given a data point $(s_t, a_t)$, we take its return $\hat{Q}(s_t, a_t)$, tokenize it, and then fine tune a model by conditioning on this return token. The gradient update is simply the log-likelihood,
\begin{equation}
\begin{aligned}
\underset{s_t, a_t \sim \mathcal{D}}{\mathbb{E}} \left[ \nabla_{\theta} \log \pi_\theta (a_t | s_t, \hat{Q}(s_t, a_t)) \right]
\end{aligned}
\end{equation}
At test time, we condition the model on the highest return $\hat{Q}_{top}$, \textit{i.e.} we sample sequences from $\pi_\theta(. | s_t, \hat{Q}(s_t, a_t) = \hat{Q}_{top} )$. 
We implement DT by quantizing the return $\hat{Q}(s_t, a_t)$ into $K$ bins, assigning a token for each bin, training a conditional model and at test time conditioning on the top bin. For binary rewards $\{0, 1\}$, this is equivalent to training a model on $r=0$ and $r=1$ tokens, and then conditioning on $r=1$ at test time.

One advantage of decision transformer over fine-tuning on top returns is that the model is trained to explicitly learn a decision boundary between different returns. However, both approaches have the theoretical drawback of requiring "trajectory coverage" \cite{brandfonbrener2022does}, \emph{i.e.} the training dataset must contain trajectories starting from the initial state $s_0$ that sees high return. This can be challenging in general because the number of data points needed increases exponentially with the length of the trajectory.

\subsection{Off-Policy Q-Learning}
A canonical approach to RL is Q-learning~\cite{watkins1992q}. We use an offline variant, Implicit Q-learning (ILQL) \cite{snell2022ilql}, as an off-policy Q-learning method architected for language models.

ILQL adds two extra heads to the pre-trained model, the action value head $Q_\theta(s_t, a_t)$ and the state value head $V_\psi(s_t)$. The state value $V_\psi(s_t)$ denotes the value of the sequence $s_t$, while the action value $Q_\theta(s_t, a_t)$ denotes the utility of a token $a_t$ given a sequence $s_t$. Hence the advantage $A(s_t,a_t) = Q_\theta(s_t, a_t) - V_\psi(s_t)$ is the utility of next token $a_t$ over any other alternate token. 

Before we describe how both heads are trained, we first note that ILQL does not explicitly train a policy. Instead, it defines an \emph{implicit} policy by taking the logits from pre-trained model $\pi_\beta$ and rescaling it by a weighted advantage: 
\begin{equation}
\begin{aligned}
\pi_\theta(a_t | s_t)  = \pi_\beta(a_t | s_t) \exp \left(\eta (Q_\theta(s_t, a_t) - V_\psi(s_t)) \right) 
\end{aligned}
\end{equation}
The loss for the  $Q_\theta(\cdot)$  head has two terms. The first is the temporal difference (TD) error coming from the Bellman equation. The second is a regularization for the policy to be close to the pre-trained policy $\pi_\beta$. The gradient update is a sum of these two terms:
\begin{align}
\begin{split}
\underset{ \substack{s_t, a_t, \\ s_{t+1} \sim \mathcal{D}}}{\mathbb{E}} &\left[ \nabla_{\theta} 
 Q_{\theta} (s_t, a_t) \underbrace{( r(s_t, a_t) + V_{\psi}(s_{t+1}) - Q_{\theta} (s_t, a_t) )}_{ \text{Temporal Difference Error}} \right]  \\
& \qquad \qquad - \alpha \underset{}{}\mathbb{E}_{s_t \sim \mathcal{D}} \nabla_\theta KL( \pi_{\beta} (.|s_t) || \pi_\theta (. | s_t) ) \, ,
\end{split}
\end{align}
\vspace{-2em}


Value head $V_\psi(s_t)$ is trained to approximate argmax of Q, \textit{i.e.} on constrained Bellman operator with expectile regression.  
\begin{equation}
\underset{ \substack{s_t, a_t \sim \mathcal{D}}}{\mathbb{E}} \nabla_\psi || Q_{\theta} (s_t, a_t) - V_{\psi}(s_t)||^{\tau} \, ,
\end{equation}
where $||u||^\tau = (\tau - \mathbbm{1}(u < 0)) u^2$ is the $\tau$ expectile. 

We \textit{improve upon original ILQL} \cite{snell2022ilql} by regularizing against logits of the pre-trained TF policy $\pi_\beta$ instead of the demonstrated data $\mathcal{D}$. This is more suited for settings where we may not have a lot of demonstrated data.


\subsection{On-Policy RL: PPO}
In addition to the offline RL approaches, we also compare against an online RL algorithm: Proximal Policy Optimization~\cite{schulman2017proximal}. PPO is a variant of a policy gradient approach that rolls out a trajectory with the current policy $\pi_\theta$ to sample $(s_t, a_t)$, estimates the advantage $A(s_t, a_t)$, and updates policy to maximize advantage while staying close to old policy $\pi_{\theta_{\text{old}}}$ The gradient update is,
\begin{equation}
\begin{aligned}
\underset{s_t, a_t \sim \pi_\theta}{\mathbb{E}} \left[ \frac {\nabla_{\theta}\pi_\theta (a_t | s_t)}{ \pi_{\theta_{\text{old}}} (a_t | s_t)} A (s_t, a_t)\right]
\end{aligned}
\end{equation}

\subsection{Comparison between Approaches}


\textbf{When is DT and Q-learning comparable?} While DT is relatively simple and faster to train, it has a more restrictive requirement of data coverage than Q-learning. Intuitively, it is unable to stitch together suboptimal trajectories that overlap into a better policy. However, for MDPs where such stitching is not possible, e.g. a tree, DT and ILQL are comparable in performance. We hypothesize that dialogue text generation belongs to this class of MDPs.



\textbf{When is DT and TF Top comparable?} While DT makes use of more data than TF Top, it does deal with a more complex function class (conditioning on returns). Intuitively, DT should expect to do better than TF Top only when the data TF Top throws away provides valuable information. If that information is already captured by  base TF model, then both DT and TF Top are likely to be similar.


\section{Experiments}
\label{evaluation}

\subsection{Experimental Setup}

\subsubsection{Task-oriented Dialogue Datasets}

We evaluate offline RL methods using three task-oriented dialogue datasets. These are relevant for dialogue systems designed for real-world applications, where users have specific goals and tasks that they want to accomplish. Each dataset consists of conversations between two speakers: one is the system or agent, and the other is the user or customer. We optimize rewards on system or agent utterances so as to emulate applications designed to assist an agent (e.g customer service representative) in providing helpful and human-like responses to customer queries and problems.

\textbf{MultiWoz 2.2~\cite{zang2020multiwoz}} is a widely used dataset created to evaluate performance of dialogue systems in multi-domain settings. It consists of over 10k conversations spanning 8 domains like hotel, train, restaurant, etc.

\textbf{Action Based Conversations Dataset (ABCD)~\cite{chen2021action}} contains customer-agent conversations where the agent's goal is to solve a customer problem. It consists of over 10k conversations spread over 55 user intents in the retail customer service domain.

\textbf{TaskMaster-3~\cite{byrne2019taskmaster}:} contains 23,789 conversations between users and a system on movie ticketing.

\subsubsection{Baselines and Metrics}
We choose a terminal binary reward \textsc{BERTClick}, which is a thresholded \textsc{BERTScore} \cite{zhang2019bertscore} with threshold value $0.6$. We select a value of $0.6$ qualitatively such the generated response is close enough and has similar meaning to human response. We evaluate on a range of automated similarity metrics shown to have a high correlation with human judgements like \textsc{BERTScore} \cite{zhang2019bertscore}, \textsc{BLEURT} \cite{sellam2020bleurt}, \textsc{METEOR} \cite{banerjee2005meteor} and \textsc{BLEU} \cite{papineni2002bleu}. We also do human evaluation on a subset of the data where we ask humans to rate similarity and relevance on a scale of 1-3. More details on the study in Appendix~\ref{appx:human_eval}.

We evaluate across following methods and baselines: \textbf{TF}, Base model trained via teacher forcing on all conversations, \textbf{TF All}, TF model fine tuned on entire offline RL Dataset, \textbf{TF Top}, TF model fine tuned only on data points with top returns $\mathcal{D}_{top}$, \textbf{DT}, Decision Transformer, \textbf{ILQL}, Off-policy Q-learning, \textbf{PPO}, Online RL via policy gradients.

We train the TF model on all the training data (stage 1), use this trained TF model to generate an offline RL dataset (stage 2), and finally fine tune different RL models on varying percentages of generated offline RL data (stage 3). More details on training setup are in Appendix~\ref{appx:experiment_details}. Since the generation step is expensive, we would like to be able to fine tune on subsets of offline RL dataset for improved efficiency and train time budgets.

For base models we study GPT2Medium\footnote{\url{https://huggingface.co/gpt2-medium}} \cite{radfordlanguage} and DistilGPT\footnote{\url{https://huggingface.co/distilgpt2}} \cite{sanh2019distilbert} which have 355M and 82M parameters, respectively. For real-time environments, models like distilGPT2 are preferable since they have low latency (order of 100 ms) to be used in dialogue settings. We use huggingface transformers library \cite{wolf2019huggingface} to implement TF Top, DT and trlx\footnote{\url{https://github.com/CarperAI/trlx}} for ILQL, PPO.


Finally, we evaluate models as both generators and rankers. For ranker, we score set of responses generated by base \textbf{TF} model and pick highest score. In our experiments, we found \textbf{ILQL} to be more effective as a ranker, as it trains a critic rather than an actor. Hence, we evaluate \textbf{ILQL} as a ranker. 

\subsection{Results and Analysis}

\begin{table*}[t]
\centering
\renewcommand{\arraystretch}{1.1}
\resizebox{0.9\textwidth}{!}{
\begin{tabular}{cccccccccccccc}
\toprule
& Algorithm & \multicolumn{2}{c}{\textsc{BERTClick}} & \multicolumn{2}{c}{\textsc{BERTScore}} & \multicolumn{2}{c}{\textsc{BLEURT}} & \multicolumn{2}{c}{\textsc{METEOR}} & \multicolumn{2}{c}{\textsc{BLEU}} & \multicolumn{2}{c}{\textsc{Perplexity}($\downarrow)$} \\ 
&  & 20\% & 80\% & 20\% & 80\% & 20\% & 80\% & 20\% & 80\% & 20\% & 80\% & 20\% & 80\%\\ 
\midrule
\parbox[t]{2mm}{\multirow{4}{*}{\rotatebox[origin=c]{90}{ABCD}}} 
 & TF & \multicolumn{2}{c}{0.276} & \multicolumn{2}{c}{0.404} & \multicolumn{2}{c}{0.571} & \multicolumn{2}{c}{0.370} & \multicolumn{2}{c}{0.135} & \multicolumn{2}{c}{36.36} \\  
 & TF All & 0.269 & 0.285 & 0.390 & 0.399 & 0.559 & 0.564 & 0.365 & 0.375 & 0.134 & 0.143 & 41.76 & 45.39 \\
 & TF Top & 0.281 & 0.307 & 0.388 & 0.420 & 0.559 & 0.576 & 0.358 & 0.382 & 0.135 & \tbcolorg 0.156 & 36.82 & 34.25 \\ 
 & DT & \tbcolorg 0.299 & \tbcolorg 0.321 & \tbcolorg 0.411 & \tbcolorg 0.429 & \tbcolorg 0.572 & \tbcolorg 0.582 & \tbcolorg  0.372 & \tbcolorg  0.391 & \tbcolorg 0.144 & 0.155 & 36.22 & 36.51\\ 
\midrule
\parbox[t]{2mm}{\multirow{4}{*}{\rotatebox[origin=c]{90}{\small MultiWoz 2.2}}}
 & TF & \multicolumn{2}{c}{0.130} & \multicolumn{2}{c}{0.366} & \multicolumn{2}{c}{0.512} & \multicolumn{2}{c}{0.312} & \multicolumn{2}{c}{0.074} & \multicolumn{2}{c}{48.97} \\
 & TF All & 0.148 & 0.163 & 0.368 & 0.376 & 0.512 & 0.519 & 0.308 & 0.313 & 0.085 & 0.082 & 42.62 & 45.83 \\
 & TF Top & 0.150 & \tbcolorg 0.179 & 0.373 &\tbcolorg  0.394 & 0.513 & 0.530 & 0.303 & 0.325 & 0.080 & \tbcolorg 0.092 & 42.84 & 41.54\\ 
 & DT & \tbcolorg 0.170 & 0.171 & \tbcolorg 0.380 & 0.392 &\tbcolorg  0.523 & \tbcolorg 0.531 & \tbcolorg 0.316 & \tbcolorg 0.331 & \tbcolorg 0.087 & 0.088 & 44.45 & 37.77\\ 
\midrule
\parbox[t]{2mm}{\multirow{4}{*}{\rotatebox[origin=c]{90}{\small TaskMaster-3}}} 
 & TF & \multicolumn{2}{c}{0.446} & \multicolumn{2}{c}{0.554} & \multicolumn{2}{c}{0.624} & \multicolumn{2}{c}{0.513} & \multicolumn{2}{c}{\tbcolorg 0.360} & \multicolumn{2}{c}{77.18} \\
 & TF All & 0.438 & 0.450 & 0.450 & 0.546 & 0.621 & 0.621 & 0.501 &  0.507  & 0.347  & 0.350 & 70.93 & 69.56\\
 & TF Top & 0.431 & 0.453 & 0.533 & 0.556 & 0.612 & 0.626 & 0.487 & 0.511  & 0.328 & 0.357 & 65.24 & 70.31\\ 
 & DT & 0.436 & \tbcolorg 0.460 & 0.548 & \tbcolorg 0.562 & 0.617 & \tbcolorg 0.630 & 0.498 &  \tbcolorg 0.514  & 0.342 & 0.359 & 69.00 & 74.67 \\ 
\midrule
\end{tabular}
}
\caption{Comparison across different methods on average metrics and dataset size with distilGPT2. 20\%, 80\% refer to percentage of the data used for fine-tuning offline RL methods. For consistency, BLEU scores are in [0, 1] unlike some papers converting them to [0, 100].}
\label{table:overall_metrics}
\end{table*}

We analyze the  results through a series of questions.

\subsubsection{Overall Performance Gains}

\textbf{Do offline RL methods improve on average over base teacher forcing model?}
Table~\ref{table:overall_metrics} presents average metrics for \textbf{TF}, \textbf{TF All}, \textbf{TF Top}, \textbf{DT} on all datasets. We see that on all datasets the offline RL methods improve the average reward (\textsc{BERTClick}) from $1.5 \%$ (TaskMaster) to $5\%$ (ABCD, MultiWoz). Offline RL methods also improve on other metrics not part of the reward, e.g. $2\%$ to $3\%$ on \textsc{METEOR} and $2\%$ (ABCD, MultiWoz) to $3\%$  on \textsc{BLEU} (ABCD). These improvements come without sacrificing perplexity (we compute perplexity with respect to off-the-shelf gpt2 model). Finally, we also note performance gains on TaskMaster are not as large as the other datasets. 

On most datasets and metrics, \textbf{DT} outperforms the other methods. The performance of \textbf{DT} over \textbf{TF Top} is consistent when fine-tuned on 20\% of the dataset vs 80\% (analyzed later in Fig.~\ref{fig:subsample_dt_vs_tf_top}). While Table~\ref{table:overall_metrics} shows only average metrics, we also look at the distribution over \textsc{BERTscore} in Fig.~\ref{fig:bert_score_hist}. We see that offline RL methods have a higher probability mass than TF on almost all \textsc{BERTScore} bins $\geq 0.6$. This is expected as $0.6$ is the threshold for \textsc{BERTClick} used as the reward function. The results show that improvements is not limited to any one bin, but across all bins.


\textbf{How does performance vary across multiple responses?}
An argument in favor of the base \textbf{TF} model might be that it's unfair to evaluate it on a single response. After all, it optimizes for recall, so with multiple responses, it should be able to reach the performance of offline RL methods.  

\begin{figure*}[!t]
\centering
\includegraphics[width=\textwidth, trim = 0cm 0.8cm 0cm 0cm]{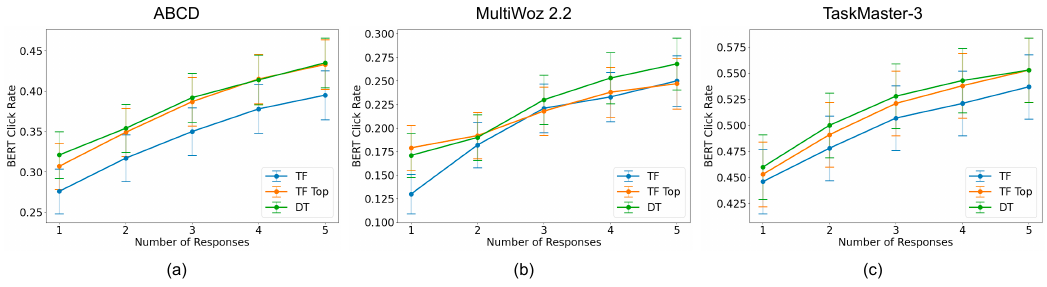}
\caption{\small Average \textsc{BERTClick} over top-k responses (w/ 95\% CI) for \textbf{(a)} ABCD \textbf{(b)} MultiWoz \textbf{(c)} Taskmaster-3 finetuned on $80\%$ data.}
\label{fig:topk_bert_click}
\figGap
\end{figure*}

Fig.~\ref{fig:topk_bert_click} shows average \textsc{BERTClick} of the best response selected from multiple responses. We see that offline RL methods maintain a persistent gap above \textbf{TF} model on all datasets. This likely indicates that they converge on a better distribution of responses over \textbf{TF}. \textbf{DT}, \textbf{TF Top} are similar for ABCD, TaskMaster, but \textbf{DT} outperforms on MultiWoz.


\begin{figure*}[!t]
\centering
\includegraphics[width=\textwidth, trim = 0cm 0.8cm 0cm 0cm]{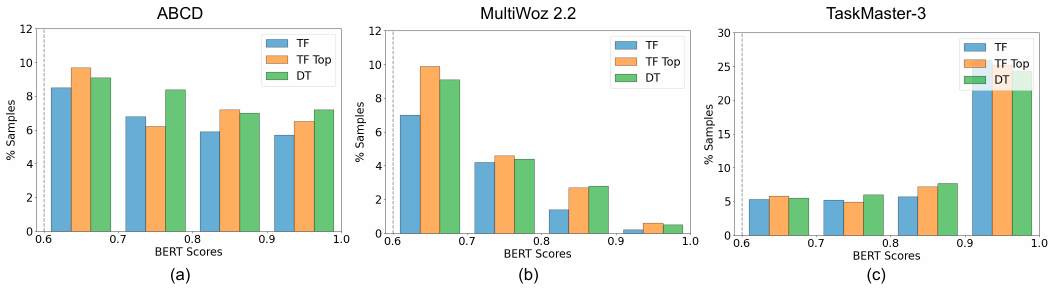}
\caption{Distribution over \textsc{BERTScores} for \textbf{(a)} ABCD \textbf{(b)} MultiWoz \textbf{(c)} Taskmaster-3 datasets with distilGPT2 finetuned on $80\%$ data.}
\vspace{-1em}
\label{fig:bert_score_hist}
\end{figure*}

\subsubsection{Human Evaluation}
\begin{figure*}[!t]
\centering
\includegraphics[width=0.96\textwidth, trim = 0cm 0.5cm 0cm 0cm]{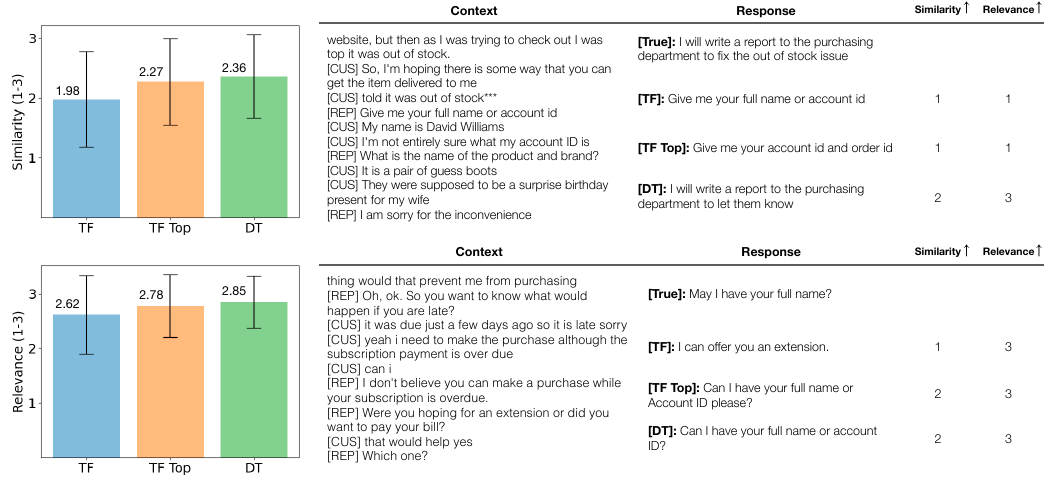}
\caption{Human evaluation (similarity and relevance) of \textbf{TF}, \textbf{TF Top}, \textbf{DT} on 100 examples with 2 representative examples presented. In the first, \textbf{DT} produces a good response while both \textbf{TF} and \textbf{TF Top} incorrectly ask the customer for their account ID even though they previously specified they do not have it. In the second, all 3 produce relevant responses though \textbf{TF} suggests something wildly dissimilar.}
\vspace{-1em}
\label{fig:human_eval}
\end{figure*}

\textbf{How do improvements look qualitatively to human evaluators?}
Fig. \ref{fig:human_eval} presents a human evaluation on 100 examples for models fine-tuned on 80\% of the data. Human evaluators were presented with a context, true human response and 3 generated responses (for each method, which are randomized and anonymized). Humans provide two ratings (1-3) -- similarity and relevance. Similarity captures how similar the response is to the true human response. Relevance captures how relevant the response is given the context (even though it may not match the human response). More details on study guidelines in Appendix~\ref{appx:human_eval}.

\textbf{DT} responses are marked the most similar ($2.36$) compared to \textbf{TF Top} ($2.27$) and \textbf{TF} ($1.98$). Interestingly, all methods do better in relevance, \textit{i.e.} \textbf{TF} ($2.62$), \textbf{TF Top} ($2.78$) and \textbf{DT} ($2.85$). This indicates while \textbf{TF} may not be producing similar utterances to humans, it is still producing relevant utterances. Offline RL fine-tunes this to prefer responses that tend to be more similar to humans.

We pick two representative example conversations in Fig. \ref{fig:human_eval}. In the first, \textbf{DT} produces both relevant and similar responses. However, both \textbf{TF} and \textbf{TF Top} produce utterances that contradict facts in the conversation, e.g, asking for account ID even though the customer said they didn't have it. 

The second example shows a case where all three methods produce relevant responses, but \textbf{TF} produces a dissimilar response, e.g. offering an extension to the customer instead of collecting personal information. More qualitative examples in Appendix~\ref{appx:qual_results}.

\textbf{How statistically significant are the improvements of TF Top and DT over TF?}
To measure statistical significance, we conduct a two sample test on the human evaluation study and provide p-values in Table~\ref{table:p_values}. While the number of examples is limited, we find improvements of both TF Top and DT over the base TF model to be statistically significant. 

\begin{table}[!h]
\centering
\resizebox{\columnwidth}{!}{
\begin{tabular}{ccccccc}
\toprule
Eval Metric & TF Top $>$ TF & DT $>$ TF \\ 
\midrule
Similarity p-value (paired t-test)   & 3.96e-03	& 2.35e-04 \\
Relevance p-value (paired t-test)  &   4.26e-02 & 4.29e-03 \\
BERTClick p-value (paired t-test)  &   7.83e-04 & 3.86e-06 \\
 \bottomrule
\end{tabular}
}
\caption{Statistical significance of human evaluation in Fig.~\ref{fig:human_eval}} \vspace{-1em}
\label{table:p_values}
\end{table}

\subsubsection{Comparison between RL methods}

\paragraph{How does ILQL critic perform as a ranker?}

Table~\ref{table:scorer_metrics} presents a comparison of all methods when ranking responses produced by the base \textbf{TF} model. \textbf{ILQL} has the largest \textsc{BERTClick} improvement of $3$\% on ABCD. It outperforms both \textbf{TF Top} (0.266) and \textbf{DT} (0.257) by a large margin. One reason for this is that  \textbf{ILQL} explicitly trains a critic $V(s)$ to approximate the optimal value.


\begin{table}[!t]
\centering
\resizebox{\columnwidth}{!}{
\begin{tabular}{ccccccc}
\toprule
 & Algorithm & \textsc{BERTClick} & \textsc{BERTScore} & \textsc{BLEURT} & \textsc{METEOR} & \textsc{BLEU} \\ 
&   & (reward) &  &  &  &  \\ 
\midrule
 \parbox[t]{2mm}{\multirow{5}{*}{\rotatebox[origin=c]{90}{ABCD}}} 
 & TF & 0.276           & 0.404           & 0.571              & 0.370               &  0.135 \\ 
   &\specialcell{DT \\ (Offline RL)}       &  \tbcolorg{0.314}           & \tbcolorg{0.425}         &  \tbcolorg{0.580}       & \tbcolorg{0.388}               & \tbcolorg{0.158} \\ 
 &\specialcell{PPO \\ (Online RL)}   &  0.274    & 0.407 & 0.578 & 0.377 & 0.143 \\
 \midrule
 \parbox[t]{2mm}{\multirow{5}{*}{\rotatebox[origin=c]{90}{\small MultiWoz 2.2}}} 
 & TF & 0.13 & 0.366 & 0.512 & 0.312 & 0.074 \\ 
   &\specialcell{DT \\ (Offline RL)}  & \tbcolorg{0.176} & \tbcolorg{0.394} & \tbcolorg{0.532} & \tbcolorg{0.334} & \tbcolorg{0.091} \\ 
 &\specialcell{PPO \\ (Online RL)}   & 0.147 & 0.364 & 0.516 & 0.320 & 0.079  \\
 \midrule
 \parbox[t]{2mm}{\multirow{5}{*}{\rotatebox[origin=c]{90}{\small TaskMaster-3}}} 
 & TF & 0.446 & 0.554 & 0.624 & 0.513 & 0.360 \\ 
   &\specialcell{DT \\ (Offline RL)}  & \tbcolorg{0.465} & \tbcolorg{0.563} & \tbcolorg{0.633} & \tbcolorg{0.521} & \tbcolorg{0.364} \\ 
 &\specialcell{PPO \\ (Online RL)}   & 0.452  & 0.561  & 0.625  & 0.510 & 0.360 \\
 \bottomrule
\end{tabular}
}
\caption{Comparison of offline RL (DT) against online RL (PPO. While \textbf{PPO} performs better than \textbf{TF}, it still performs worse than \textbf{DT} on all datasets.} \vspace{-1em}
\label{table:ppo_comparison}
\end{table}

\begin{table}[!t]
\centering
\resizebox{\columnwidth}{!}{
\begin{tabular}{ccccccc}
\toprule
 & Algorithm & \textsc{BERTClick} & \textsc{BERTScore} & \textsc{BLEURT} & \textsc{METEOR} & \textsc{BLEU} \\ 
&   & (reward) &  &  &  &  \\ 
\midrule
 \parbox[t]{2mm}{\multirow{5}{*}{\rotatebox[origin=c]{90}{ABCD}}} 
 & TF       & 0.251     & 0.387 & 0.571 & 0.383 & 0.13\\ 
 & TF All      & 0.244     & 0.377 & 0.566 & 0.385 & 0.13 \\
 & TF Top   & 0.266    & 0.398 & \tbcolorg 0.572 & \tbcolorg 0.399 & 0.13 \\ 
 & DT       & 0.257     & 0.388 & 0.570 & 0.392 & 0.12 \\ 
 & ILQL     & \tbcolorg 0.285     & \tbcolorg 0.403 & 0.568 & 0.366 & \tbcolorg 0.14 \\
 \midrule
 \parbox[t]{2mm}{\multirow{5}{*}{\rotatebox[origin=c]{90}{Taskmaster-3}}} 
 & TF     &  0.388  &  0.49  & 0.584 & 0.485 & 0.296 \\ 
 & TF All & 0.377 & 0.477 & 0.58 & 0.478  & 0.277 \\ 
 & TF Top & 0.426  &  0.512  &  0.598  & 0.499 & 0.303 \\ 
 & DT    & \tbcolorg 0.442  &  0.512   &  \tbcolorg 0.597  & \tbcolorg 0.496 & 0.297  \\ 
 & ILQL   & 0.439  &  \tbcolorg 0.52  & 0.593  & 0.486 & \tbcolorg 0.306 \\ 
 \bottomrule
\end{tabular}
}
\caption{Comparison when ranking responses generated by the base TF model. Offline RL methods improve over logit scoring of base TF model, with \textbf{ILQL} being most effective as a ranker.} \vspace{-1em}
\label{table:scorer_metrics}
\end{table}

\textbf{How do offline RL compare with PPO?}
Table~\ref{table:ppo_comparison} presents a comparison of \textbf{PPO} against \textbf{DT} and \textbf{TF}. While \textbf{PPO} performs better than \textbf{TF}, it still performs worse than \textbf{DT} on all datasets. During training, \textbf{PPO} reward for the model over iterations appear unstable: 0.272 (epoch=1), 0.268 (epoch=3), 0.274 (epoch=5). \textbf{DT} on the other hand shows much more stable convergence. This is consistent with the discussion in \ref{sec:why_offline_rl} that for text generation, on-policy exploration can be challenging and requires significant KL regularization to the base \textbf{TF} policy. This KL regularization serves to limit the performance gains. We see the following trend for average reward: 0.259 (KL=0.1), 0.274 (KL=0.2), 0.279 (KL=0.4). For very high KL, performance falls back to the base TF reward of 0.276. \textbf{PPO} also has much longer training times  because of calls it has to make to the model’s generate function and \textsc{BERTScore} computation. \textbf{PPO} takes 1.95 hours / epoch, while \textbf{DT} takes 1.24 hours / epoch and \textbf{TF Top} takes 0.48 hours / epoch. 

\textbf{Can online data collection help DT?}
We compare with \textbf{Quark}~\cite{lu2022quark}, which can be viewed as an online counterpart to \textbf{DT}. It introduces an outer loop on \textbf{DT} by iteratively training a model, collecting data with the model and retraining. While this requires an extra outer loop for collecting data, this can certainly improve performance by collecting more positive examples on-policy as the policy improves. We implement \textbf{Quark} by creating an outer loop where at every epoch we collect new data with the current policy. We compare this to \textbf{DT} that holds the data fixed across epochs. 

\begin{figure}[!h]
\centering
\includegraphics[width=0.8\columnwidth, trim = 0cm 0.8cm 0cm 0cm]{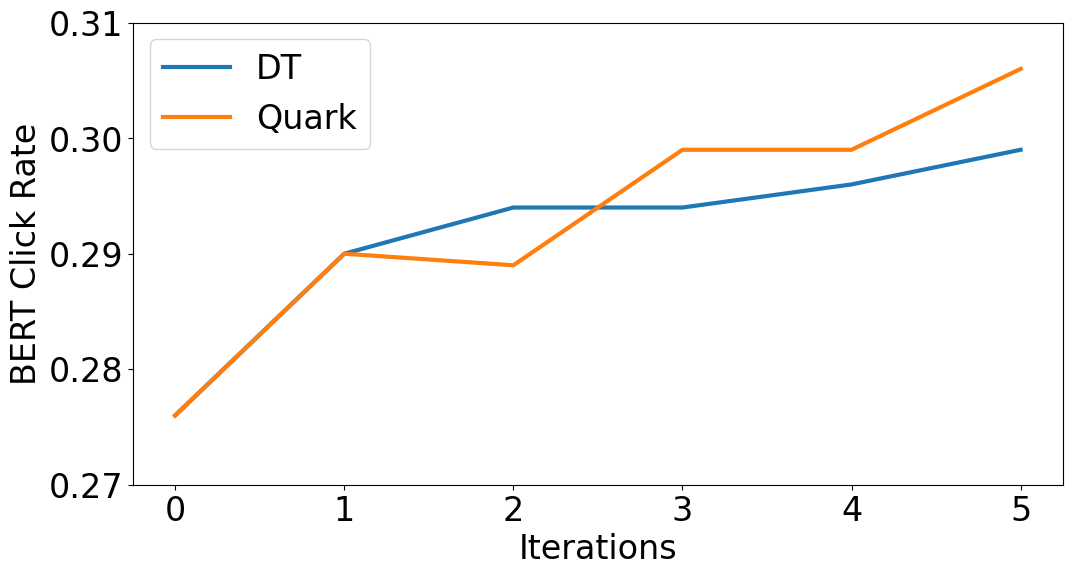}
\caption{Average \textsc{BERTClick} for \textbf{DT} vs \textbf{Quark}}
\label{fig:dt_vs_quark}
\end{figure}

Fig.~\ref{fig:dt_vs_quark} shows \textsc{BERTClick} of \textbf{DT} and \textbf{Quark} over iterations fine-tuned on $20\%$ ABCD dataset. While performance is comparable in the initial epochs, the online data collection seems to help \textbf{Quark} outperform \textbf{DT} at the end of epoch 5. However, the performance boost with an additional online data collection step would vary with tasks depending on how good a coverage sampling from the base TF model has.



\begin{table}[t]
\centering
\resizebox{\columnwidth}{!}{
\begin{tabular}{cccccccc}
\toprule
& & Algorithm & \textsc{BERTClick} & \textsc{BERTScore} & \textsc{BLEURT} & \textsc{METEOR} & \textsc{BLEU} \\ 
 & &  & (reward) &  &  &  \\ 
\midrule
\parbox[t]{2mm}{\multirow{4}{*}{\rotatebox[origin=c]{90}{distilGPT2}}} &
\parbox[t]{2mm}{\multirow{4}{*}{\rotatebox[origin=c]{90}{\tiny ~\cite{sanh2019distilbert}}}}
 & TF   & 0.276    &   0.404    &   0.571     &  0.37  & 0.135  \\
 && TF All & 0.285    &   0.399    &    0.564    & 0.375  & 0.143 \\
 && TF Top & 0.307    &    0.42    &    0.576    & 0.382  & \tbcolorg 0.156 \\ 
 && DT    & \tbcolorg 0.321    &   \tbcolorg 0.429   &   \tbcolorg 0.582    & \tbcolorg 0.391  & 0.155 \\ 
 \midrule
 \parbox[t]{2mm}{\multirow{4}{*}{\rotatebox[origin=c]{90}{GPT2 Med}}} &
 \parbox[t]{2mm}{\multirow{4}{*}{\rotatebox[origin=c]{90}{\tiny 
\scalebox{.9}{~\cite{radfordlanguage}}}}} 
 & TF  &  0.278   &   0.414    &   0.577   & 0.369  & 0.139 \\ 
 && TF All & 0.309    &   0.422   &    0.581  & 0.39  & 0.157 \\ 
 && TF Top &  0.331   &  0.444    &    0.596  & \tbcolorg 0.407  & 0.162  \\ 
 && DT  & \tbcolorg 0.334   &   \tbcolorg 0.446    &   \tbcolorg 0.597  & 0.406  & \tbcolorg 0.163  \\
 \bottomrule
\end{tabular}
}
\caption{Comparison across different model sizes. Improvements are continually sustained as we go to a larger model size.} \vspace{-1.em}
\label{table:model_sizes}
\end{table}

\subsubsection{Ablations and Analysis}

\paragraph{How do offline RL improvements vary with model size?}

As we increase the model size from distilGPT2 to GPT2 Med, we see performance of all methods improves. However, offline RL methods persistently maintain a $5$\% performance gain over \textbf{TF} across sizes. This indicates that offline RL performance gains come from the way the model is trained rather than simply having a larger model capacity. 
 
\begin{figure}[!t]
\centering
\includegraphics[width=\columnwidth, trim = 0cm 0.8cm 0cm 0cm]{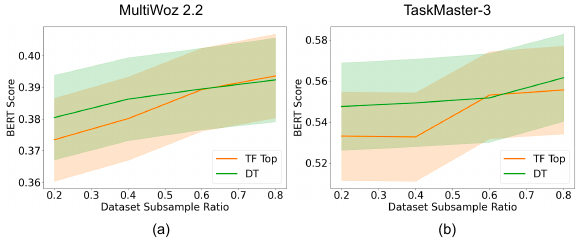}
\caption{TF Top vs DT performance with increasing offline RL data size on \textbf{(a)} MultiWoz 2.2 and \textbf{(b)} Taskmaster-3 datasets. DT significantly outperforms TF Top with limited data, with performance gap narrowing with more data.}
\label{fig:subsample_dt_vs_tf_top}
\figGap
\end{figure}

\paragraph{How does performance vary with offline RL data size?}
\label{sec:subsample_dt_vs_tf_top}
Fig. \ref{fig:subsample_dt_vs_tf_top} shows how performance of offline RL varies with increasing data. \textbf{DT} has an edge at low data size, but as data size increases \textbf{TF Top} and \textbf{DT} merge. This backs our understanding from the theory behind \textbf{DT} and \textbf{TF Top}, where \textbf{TF Top} throws away data while \textbf{DT} retains it. This advantage goes away with increasing data size. It's important to note that for fine-tuning, we will often be in the low data regime and hence \textbf{DT} is favourable from that regard. 

\paragraph{How does TF Top performance vary with different top quantiles?}

\begin{figure}[!t]
\centering
\includegraphics[width=\columnwidth, trim = 0cm 0.8cm 0cm 0cm]{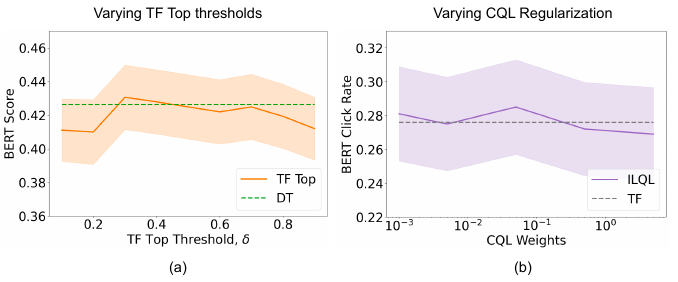}
\caption{ABCD dataset ablations for (a) Average \textsc{BERTScore} (reward) with varying TF Top thresholds on what constitutes as top returns. \textbf{(b)} Average \textsc{BERTClick} (reward) for ILQL with varying regularization against base TF model logits.} \vspace{-1em}
\label{fig:ablation_tf_top_thresh_and_cql_reg}
\end{figure}

We conducted an ablation experiment where we trained both \textbf{DT} and \textbf{TF Top} with varying \textsc{BERTClick} thresholds. Fig. \ref{fig:ablation_tf_top_thresh_and_cql_reg} shows the average \textsc{BERTScore} of the greedy response.

As we increase the quantile threshold, we see the \textbf{TF Top} performance increase, reach a peak and then drop. On one extreme, setting the threshold to be $0$ implies that we are training \textbf{TF Top} on all the data. This is suboptimal as \textbf{TF Top} trains on all of it's own responses and fails to tell the difference between good and bad responses. On the other extreme, setting the threshold to be $1$ implies that we are training \textbf{TF Top} on only the human response, which has similar performance to \textbf{TF}. 

\paragraph{How does ILQL performance change with varying regularization?}

As we increase regularization  $\alpha$, \textbf{ILQL} performance improves as it forces the critic to stay close to the data. Increasing $\alpha$ further ($>0.05$ in Fig. \ref{fig:ablation_tf_top_thresh_and_cql_reg}(b)) hurts performance as the regularization dominates other losses.  

\paragraph{How does offline RL compare with TF on dialogue metrics?}
We analyze how various methods perform on dialogue metrics, i.e., metrics looking at whether the generated response results in a correct slot prediction. We chose a state-of-the-art approach~\cite{lee2021dialogue} to train a T5 dialogue state tracking (DST) model on MultiWoz to extract slots from generated responses. 

In MultiWoz, We took generated responses from the offline RL method and replaced “SYSTEM” utterances with the generated responses (keeping “USER” utterances the same). We then feed these to the DST model and compute different dialogue-level joint accuracy metrics: 'joint\_goal\_accuracy', 'joint\_cat\_accuracy', 'joint\_noncat\_accuracy’ in Table~\ref{table:dialogue_metrics}. 

\begin{table}[!h]
\centering
\resizebox{\columnwidth}{!}{
\begin{tabular}{ccccccc}
\toprule
 & \textbf{joint\_goal\_accuracy} & \textbf{joint\_cat\_accuracy} & \textbf{joint\_noncat\_accuracy} \\ 
\midrule
\textbf{Groundtruth}  &   0.565	 & 0.712 &  0.766\\
\textbf{TF Top}    & 0.474	 & 0.689 & 0.629\\
\textbf{TF} & 0.458  & 0.679 & 0.613\\
 \bottomrule
\end{tabular}
}
\caption{Dialogue metrics on MultiWoz dataset} \vspace{-1.em}
\label{table:dialogue_metrics}
\end{table}

Overall, we find that both \textbf{TF Top} and \textbf{TF} do worse than the ground truth, as expected. Ground truth utterances have access to privileged information which in turn defines the ground truth slots. For instance, a specific restaurant name that neither of the generated utterances would be able to predict ahead of time. 
Interestingly, we see \textbf{TF Top} score higher than \textbf{TF} on the slot metrics even though such metrics do not appear in the rewards. When looking at the utterances, we observe that TF makes mistakes by either making up new information or repeating information from the context (similar to the qualitative / human study examples in the paper). However, for offline RL methods to truly do better on these metrics, they must be trained on rewards that capture such dialogue-level metrics. This is an interesting direction for future work.







\section{Discussion}
\label{discussion}

In this paper, we examine the effectiveness of offline RL methods for generating dialogue responses. We analyze three distinct techniques: fine-tuning on high returns (TF Top), conditioning on return (DT), and an off-policy Q-learning approach (ILQL). Our evaluation is based on three task-oriented dialogue datasets, and we conduct various analyses and ablation studies to investigate the trade-offs between these approaches. 

\paragraph{Offline RL models learn to produce good enough text that are similar to human.}


We hypothesized that there are multiple ways to convey the same information as a human and that a model can learn this. We constructed a reward using \textsc{BERTscore} that captures this similarity and trained various offline RL methods. 
Our results show that offline RL clearly improves upon traditional methods by approximately 5\% (Table \ref{table:overall_metrics}). We found that the improvements were most significant in examples where traditional methods repeat themselves to ask for the same information or do not follow the correct flow of a human utterance even if the response is contextually relevant (Fig. \ref{fig:human_eval}). Improvements were not limited to overall averages but also seen as a distributional improvement (Fig. \ref{fig:bert_score_hist}). Additionally, the improvements were sustained across multiple responses and when using larger models (Fig. \ref{fig:topk_bert_click}, Table \ref{table:model_sizes}).

\paragraph{Decision Transformer is a practical choice.}

When working with all available data, \textbf{DT} and \textbf{TF Top} show comparable performance. However, when it comes to limited data, \textbf{DT} significantly outperforms \textbf{TF Top} (Table \ref{table:overall_metrics}, Fig. \ref{fig:subsample_dt_vs_tf_top}). This aligns with our understanding from theory that suggests that \textbf{TF Top} discards useful information while \textbf{DT} retains it. This is relevant for fine-tuning in low data regimes where we expect \textbf{DT} to be more effective. 


\paragraph{We see two potential future directions.} First, we use \textsc{BERTscore} as a proxy for whether a human would have clicked on the suggested utterance. Instead, can we learn reward functions from human feedback that is easier to optimize? Second, we consider a single turn when a dialogue has multiple turns. How do these methods compare when optimizing rewards that extend to more than 1 turn? 



\small
\section*{Acknowledgements}
We would like to thank Yoav Artzi and the entire ASAPP research team for their insightful feedback and suggestions. We greatly appreciate efforts from the in-house data curation team, in particular Nick, Molly, Kyle, Divya, in providing high-quality annotations. Lastly, we are grateful to the anonymous reviewers, whose detailed feedback and constructive suggestions significantly improved the quality of this paper.


\balance
\bibliography{references}

\begin{thebibliography}{46}
\providecommand{\natexlab}[1]{#1}
\providecommand{\url}[1]{\texttt{#1}}
\expandafter\ifx\csname urlstyle\endcsname\relax
  \providecommand{\doi}[1]{doi: #1}\else
  \providecommand{\doi}{doi: \begingroup \urlstyle{rm}\Url}\fi

\bibitem[Banerjee \& Lavie(2005)Banerjee and Lavie]{banerjee2005meteor}
Banerjee, S. and Lavie, A.
\newblock {METEOR}: An automatic metric for {MT} evaluation with improved
  correlation with human judgments.
\newblock In \emph{Proceedings of the {ACL} Workshop on Intrinsic and Extrinsic
  Evaluation Measures for Machine Translation and/or Summarization}, pp.\
  65--72, Ann Arbor, Michigan, June 2005. Association for Computational
  Linguistics.
\newblock URL \url{https://aclanthology.org/W05-0909}.

\bibitem[Bang et~al.(2023)Bang, Lee, and Koo]{bang2023task}
Bang, N., Lee, J., and Koo, M.-W.
\newblock Task-optimized adapters for an end-to-end task-oriented dialogue
  system.
\newblock \emph{arXiv preprint arXiv:2305.02468}, 2023.

\bibitem[Brandfonbrener et~al.(2022)Brandfonbrener, Bietti, Buckman, Laroche,
  and Bruna]{brandfonbrener2022does}
Brandfonbrener, D., Bietti, A., Buckman, J., Laroche, R., and Bruna, J.
\newblock When does return-conditioned supervised learning work for offline
  reinforcement learning?
\newblock In \emph{Advances in Neural Information Processing Systems
  (NeurIPS)}, 2022.

\bibitem[Byrne et~al.(2019)Byrne, Krishnamoorthi, Sankar, Neelakantan,
  Duckworth, Yavuz, Goodrich, Dubey, Cedilnik, and Kim]{byrne2019taskmaster}
Byrne, B., Krishnamoorthi, K., Sankar, C., Neelakantan, A., Duckworth, D.,
  Yavuz, S., Goodrich, B., Dubey, A., Cedilnik, A., and Kim, K.-Y.
\newblock Taskmaster-1: Toward a realistic and diverse dialog dataset.
\newblock In \emph{Empirical Methods in Natural Language Processing (EMNLP)},
  2019.

\bibitem[Chen et~al.(2021{\natexlab{a}})Chen, Chen, Yang, Lin, and
  Yu]{chen2021action}
Chen, D., Chen, H., Yang, Y., Lin, A., and Yu, Z.
\newblock Action-based conversations dataset: A corpus for building more
  in-depth task-oriented dialogue systems.
\newblock In \emph{Proc. North American Chapter of the Association for
  Computational Linguistics (NAACL)}, pp.\  3002--3017, 2021{\natexlab{a}}.

\bibitem[Chen et~al.(2021{\natexlab{b}})Chen, Lu, Rajeswaran, Lee, Grover,
  Laskin, Abbeel, Srinivas, and Mordatch]{chen2021decision}
Chen, L., Lu, K., Rajeswaran, A., Lee, K., Grover, A., Laskin, M., Abbeel, P.,
  Srinivas, A., and Mordatch, I.
\newblock Decision transformer: Reinforcement learning via sequence modeling.
\newblock In \emph{Advances in Neural Information Processing Systems
  (NeurIPS)}, 2021{\natexlab{b}}.

\bibitem[Furman et~al.(2022)Furman, Toledo, Shock, and Buys]{furman2022qait}
Furman, G., Toledo, E., Shock, J., and Buys, J.
\newblock A sequence modelling approach to question answering in text-based
  games.
\newblock In \emph{ACL Wordplay Workshop}, 2022.

\bibitem[Jaques et~al.(2019)Jaques, Ghandeharioun, Shen, Ferguson, Lapedriza,
  Jones, Gu, and Picard]{jaques2019wayoff}
Jaques, N., Ghandeharioun, A., Shen, J.~H., Ferguson, C., Lapedriza, A., Jones,
  N., Gu, S., and Picard, R.
\newblock Way off-policy batch deep reinforcement learning of implicit human
  preferences in dialog.
\newblock \emph{arXiv preprint arXiv:1907.00456}, 2019.

\bibitem[Jaques et~al.(2020)Jaques, Shen, Ghandeharioun, Ferguson, Lapedriza,
  Jones, Gu, and Picard]{jaques2020human}
Jaques, N., Shen, J.~H., Ghandeharioun, A., Ferguson, C., Lapedriza, A., Jones,
  N., Gu, S.~S., and Picard, R.
\newblock Human-centric dialog training via offline reinforcement learning.
\newblock \emph{arXiv preprint arXiv:2010.05848}, 2020.

\bibitem[Kiegeland \& Kreutzer(2021)Kiegeland and
  Kreutzer]{kiegeland2021revisiting}
Kiegeland, S. and Kreutzer, J.
\newblock Revisiting the weaknesses of reinforcement learning for neural
  machine translation.
\newblock \emph{arXiv preprint arXiv:2106.08942}, 2021.

\bibitem[Kostrikov et~al.(2021)Kostrikov, Nair, and
  Levine]{kostrikov2021offline}
Kostrikov, I., Nair, A., and Levine, S.
\newblock Offline reinforcement learning with implicit q-learning.
\newblock \emph{arXiv preprint arXiv:2110.06169}, 2021.

\bibitem[Kumar et~al.(2020)Kumar, Zhou, Tucker, and
  Levine]{kumar2020conservative}
Kumar, A., Zhou, A., Tucker, G., and Levine, S.
\newblock Conservative q-learning for offline reinforcement learning.
\newblock \emph{Advances in Neural Information Processing Systems},
  33:\penalty0 1179--1191, 2020.

\bibitem[Lee et~al.(2021)Lee, Cheng, and Ostendorf]{lee2021dialogue}
Lee, C.-H., Cheng, H., and Ostendorf, M.
\newblock Dialogue state tracking with a language model using schema-driven
  prompting.
\newblock \emph{arXiv preprint arXiv:2109.07506}, 2021.

\bibitem[Levine et~al.(2020)Levine, Kumar, Tucker, and Fu]{levine2020offline}
Levine, S., Kumar, A., Tucker, G., and Fu, J.
\newblock Offline reinforcement learning: Tutorial, review, and perspectives on
  open problems.
\newblock \emph{arXiv preprint arXiv:2005.01643}, 2020.

\bibitem[Li et~al.(2016)Li, Monroe, Ritter, Galley, Gao, and
  Jurafsky]{li2016deep}
Li, J., Monroe, W., Ritter, A., Galley, M., Gao, J., and Jurafsky, D.
\newblock Deep reinforcement learning for dialogue generation.
\newblock \emph{arXiv preprint arXiv:1606.01541}, 2016.

\bibitem[Liu et~al.(2016)Liu, Lowe, Serban, Noseworthy, Charlin, and
  Pineau]{liu2016not}
Liu, C.-W., Lowe, R., Serban, I.~V., Noseworthy, M., Charlin, L., and Pineau,
  J.
\newblock How not to evaluate your dialogue system: An empirical study of
  unsupervised evaluation metrics for dialogue response generation.
\newblock \emph{arXiv preprint arXiv:1603.08023}, 2016.

\bibitem[Lu et~al.(2022)Lu, Welleck, Jiang, Hessel, Qin, West, Ammanabrolu, and
  Choi]{lu2022quark}
Lu, X., Welleck, S., Jiang, L., Hessel, J., Qin, L., West, P., Ammanabrolu, P.,
  and Choi, Y.
\newblock Quark: Controllable text generation with reinforced unlearning.
\newblock In \emph{Advances in Neural Information Processing Systems
  (NeurIPS)}, 2022.

\bibitem[Misra \& Artzi(2015)Misra and Artzi]{misra2015reinforcement}
Misra, D. and Artzi, Y.
\newblock Reinforcement learning for mapping instructions to actions with
  reward learning.
\newblock 2015.

\bibitem[Ouyang et~al.(2022)Ouyang, Wu, Jiang, Almeida, Wainwright, Mishkin,
  Zhang, Agarwal, Slama, Ray, et~al.]{ouyang2022training}
Ouyang, L., Wu, J., Jiang, X., Almeida, D., Wainwright, C.~L., Mishkin, P.,
  Zhang, C., Agarwal, S., Slama, K., Ray, A., et~al.
\newblock Training language models to follow instructions with human feedback.
\newblock \emph{arXiv preprint arXiv:2203.02155}, 2022.

\bibitem[Pang \& He(2020)Pang and He]{pang2020text}
Pang, R.~Y. and He, H.
\newblock Text generation by learning from demonstrations.
\newblock In \emph{International Conference on Learning Representations
  (ICLR)}, 2020.

\bibitem[Papineni et~al.(2002)Papineni, Roukos, Ward, and
  Zhu]{papineni2002bleu}
Papineni, K., Roukos, S., Ward, T., and Zhu, W.-J.
\newblock Bleu: a method for automatic evaluation of machine translation.
\newblock In \emph{Proceedings of the 40th annual meeting of the Association
  for Computational Linguistics}, pp.\  311--318, 2002.

\bibitem[Pasunuru \& Bansal(2018)Pasunuru and Bansal]{pasunuru2018multi}
Pasunuru, R. and Bansal, M.
\newblock Multi-reward reinforced summarization with saliency and entailment.
\newblock \emph{arXiv preprint arXiv:1804.06451}, 2018.

\bibitem[Paulus et~al.(2017)Paulus, Xiong, and Socher]{paulus2017deep}
Paulus, R., Xiong, C., and Socher, R.
\newblock A deep reinforced model for abstractive summarization.
\newblock \emph{arXiv preprint arXiv:1705.04304}, 2017.

\bibitem[Radford et~al.(2019)Radford, Wu, Child, Luan, Amodei, and
  Sutskever]{radfordlanguage}
Radford, A., Wu, J., Child, R., Luan, D., Amodei, D., and Sutskever, I.
\newblock Language models are unsupervised multitask learners.
\newblock 2019.

\bibitem[Ramakrishnan et~al.(2022)Ramakrishnan, Narangodage, Schilman,
  Weinberger, and McDonald]{ramakrishnan2022long}
Ramakrishnan, R., Narangodage, H.~B., Schilman, M., Weinberger, K.~Q., and
  McDonald, R.
\newblock Long-term control for dialogue generation: Methods and evaluation.
\newblock In \emph{Proc. North American Chapter of the Association for
  Computational Linguistics (NAACL)}, 2022.

\bibitem[Ramamurthy et~al.(2022)Ramamurthy, Ammanabrolu, Brantley, Hessel,
  Sifa, Bauckhage, Hajishirzi, and Choi]{ramamurthy2022reinforcement}
Ramamurthy, R., Ammanabrolu, P., Brantley, K., Hessel, J., Sifa, R., Bauckhage,
  C., Hajishirzi, H., and Choi, Y.
\newblock Is reinforcement learning (not) for natural language processing?:
  Benchmarks, baselines, and building blocks for natural language policy
  optimization.
\newblock \emph{arXiv preprint arXiv:2210.01241}, 2022.

\bibitem[Ranzato et~al.(2015)Ranzato, Chopra, Auli, and
  Zaremba]{ranzato2015sequence}
Ranzato, M., Chopra, S., Auli, M., and Zaremba, W.
\newblock Sequence level training with recurrent neural networks.
\newblock In \emph{International Conference on Learning Representations
  (ICLR)}, 2015.

\bibitem[Sanh et~al.(2019)Sanh, Debut, Chaumond, and Wolf]{sanh2019distilbert}
Sanh, V., Debut, L., Chaumond, J., and Wolf, T.
\newblock Distilbert, a distilled version of bert: smaller, faster, cheaper and
  lighter.
\newblock In \emph{NeurIPS EMC\^2 Workshop}, 2019.

\bibitem[Schulman et~al.(2017)Schulman, Wolski, Dhariwal, Radford, and
  Klimov]{schulman2017proximal}
Schulman, J., Wolski, F., Dhariwal, P., Radford, A., and Klimov, O.
\newblock Proximal policy optimization algorithms.
\newblock \emph{arXiv preprint arXiv:1707.06347}, 2017.

\bibitem[Sellam et~al.(2020)Sellam, Das, and Parikh]{sellam2020bleurt}
Sellam, T., Das, D., and Parikh, A.~P.
\newblock Bleurt: Learning robust metrics for text generation.
\newblock \emph{arXiv preprint arXiv:2004.04696}, 2020.

\bibitem[Snell et~al.(2022)Snell, Kostrikov, Su, Yang, and
  Levine]{snell2022ilql}
Snell, C., Kostrikov, I., Su, Y., Yang, M., and Levine, S.
\newblock Offline {RL} for natural language generation with implicit language
  {Q} learning.
\newblock \emph{arXiv preprint arXiv:2206.11871}, 2022.

\bibitem[Stiennon et~al.(2020)Stiennon, Ouyang, Wu, Ziegler, Lowe, Voss,
  Radford, Amodei, and Christiano]{stiennon2020learning}
Stiennon, N., Ouyang, L., Wu, J., Ziegler, D., Lowe, R., Voss, C., Radford, A.,
  Amodei, D., and Christiano, P.~F.
\newblock Learning to summarize with human feedback.
\newblock \emph{Advances in Neural Information Processing Systems},
  33:\penalty0 3008--3021, 2020.

\bibitem[Swanson et~al.(2019)Swanson, Yu, Fox, Wohlwend, and
  Lei]{swanson2019building}
Swanson, K., Yu, L., Fox, C., Wohlwend, J., and Lei, T.
\newblock Building a production model for retrieval-based chatbots.
\newblock \emph{arXiv preprint arXiv:1906.03209}, 2019.

\bibitem[Tian et~al.(2021)Tian, Huang, Lin, Bao, He, Yang, Wu, Wang, and
  Sun]{tian2021amendable}
Tian, X., Huang, L., Lin, Y., Bao, S., He, H., Yang, Y., Wu, H., Wang, F., and
  Sun, S.
\newblock Amendable generation for dialogue state tracking.
\newblock \emph{arXiv preprint arXiv:2110.15659}, 2021.

\bibitem[Verma et~al.(2022)Verma, Fu, Yang, and Levine]{verma2022chai}
Verma, S., Fu, J., Yang, M., and Levine, S.
\newblock Chai: A chatbot ai for task-oriented dialogue with offline
  reinforcement learning.
\newblock \emph{arXiv preprint arXiv:2204.08426}, 2022.

\bibitem[Watkins \& Dayan(1992)Watkins and Dayan]{watkins1992q}
Watkins, C.~J. and Dayan, P.
\newblock Q-learning.
\newblock \emph{Machine learning}, 8\penalty0 (3):\penalty0 279--292, 1992.

\bibitem[Williams(1992)]{williams1992simple}
Williams, R.~J.
\newblock Simple statistical gradient-following algorithms for connectionist
  reinforcement learning.
\newblock \emph{Machine learning}, 8\penalty0 (3):\penalty0 229--256, 1992.

\bibitem[Williams \& Zipser(1989)Williams and Zipser]{williams1989learning}
Williams, R.~J. and Zipser, D.
\newblock A learning algorithm for continually running fully recurrent neural
  networks.
\newblock \emph{Neural computation}, 1\penalty0 (2):\penalty0 270--280, 1989.

\bibitem[Wolf et~al.(2019)Wolf, Debut, Sanh, Chaumond, Delangue, Moi, Cistac,
  Rault, Louf, Funtowicz, et~al.]{wolf2019huggingface}
Wolf, T., Debut, L., Sanh, V., Chaumond, J., Delangue, C., Moi, A., Cistac, P.,
  Rault, T., Louf, R., Funtowicz, M., et~al.
\newblock Huggingface's transformers: State-of-the-art natural language
  processing.
\newblock \emph{arXiv preprint arXiv:1910.03771}, 2019.

\bibitem[Wu et~al.(2022)Wu, Brantley, Kojima, and Artzi]{wu2022lilgym}
Wu, A., Brantley, K., Kojima, N., and Artzi, Y.
\newblock lilgym: Natural language visual reasoning with reinforcement
  learning.
\newblock \emph{arXiv preprint arXiv:2211.01994}, 2022.

\bibitem[Wu et~al.(2018)Wu, Xia, Tian, Zhao, Qin, Lai, and
  Liu]{wu2018adversarial}
Wu, L., Xia, Y., Tian, F., Zhao, L., Qin, T., Lai, J., and Liu, T.-Y.
\newblock Adversarial neural machine translation.
\newblock In \emph{Asian Conference on Machine Learning}, pp.\  534--549. PMLR,
  2018.

\bibitem[Yang et~al.(2020)Yang, Chen, and Narasimhan]{yang2020improving}
Yang, R., Chen, J., and Narasimhan, K.
\newblock Improving dialog systems for negotiation with personality modeling.
\newblock \emph{arXiv preprint arXiv:2010.09954}, 2020.

\bibitem[Yonghui et~al.(2016)Yonghui, Schuster, Chen, Le, Norouzi, Macherey,
  Krikun, Cao, Gao, Macherey, et~al.]{yonghui2016bridging}
Yonghui, W., Schuster, M., Chen, Z., Le, Q.~V., Norouzi, M., Macherey, W.,
  Krikun, M., Cao, Y., Gao, Q., Macherey, K., et~al.
\newblock Bridging the gap between human and machine translation.
\newblock \emph{arXiv preprint arXiv:1609.08144}, 2016.

\bibitem[Zang et~al.(2020)Zang, Rastogi, Sunkara, Gupta, Zhang, and
  Chen]{zang2020multiwoz}
Zang, X., Rastogi, A., Sunkara, S., Gupta, R., Zhang, J., and Chen, J.
\newblock {M}ulti{WOZ} 2.2 : A dialogue dataset with additional annotation
  corrections and state tracking baselines.
\newblock In \emph{Proceedings of the 2nd Workshop on Natural Language
  Processing for Conversational AI}, 2020.
\newblock URL \url{https://aclanthology.org/2020.nlp4convai-1.13}.

\bibitem[Zhang et~al.(2019)Zhang, Kishore, Wu, Weinberger, and
  Artzi]{zhang2019bertscore}
Zhang, T., Kishore, V., Wu, F., Weinberger, K.~Q., and Artzi, Y.
\newblock {BERT}score: Evaluating text generation with {BERT}.
\newblock \emph{arXiv preprint arXiv:1904.09675}, 2019.

\bibitem[Zhou et~al.(2017)Zhou, Small, Rokhlenko, and Elkan]{zhou2017end}
Zhou, L., Small, K., Rokhlenko, O., and Elkan, C.
\newblock End-to-end offline goal-oriented dialog policy learning via policy
  gradient.
\newblock \emph{arXiv preprint arXiv:1712.02838}, 2017.

\end{thebibliography}
\bibliographystyle{icml2023}


\appendix
\onecolumn

\section{Limitations}
In this paper, we study small and medium size models due to the limited computation resources. It is possible that our findings do not generalize to large scale models with billions of parameters. We plan to extend our study to large language models in a future work.

\section{Potential Negative Social Impacts}
It is possible that dialogue language models are used to generate malicious text or produce harmful conversations with humans if they are trained on biased data. Directly applying a language model to a product without any safeguard is risky. The models trained and released with this paper should not be used in any products.

\section{Human Evaluation Details}
\label{appx:human_eval}

We gather evaluations from humans to assess the quality of response utterances generated given a conversational context. We use two measures to evaluate the generated utterances: (a) how similar they are to the actual response, and (b) how relevant they are to the context. We obtain annotations from 5 different annotators on 100 examples from ABCD dataset, each annotator evaluating 20 examples. We provided the following guidelines to the annotators:

\subsection{Similarity to True Response?}

On a scale of 1 to 3 (1=not similar, 3=similar), how similar is the generated response to the true response?

\subsubsection*{Scale}

\textbf{1 = not similar}

Not similar at all or even opposite in meaning. This can even include sentences that have a lot of string overlap, e.g., "I booked that for you", "I didn’t book that for you".

\textbf{2 = somewhat similar}

Overlap in meaning, but some errors or missing / added information. E.g., ‘You need to bring your passport’ and ‘You need to bring your passport and vaccination record’ or ‘You need to bring your passport’ and ‘You need to bring your identity card’

\textbf{3 = similar}

Essentially the same meaning. A human reading the two responses would come to basically the same conclusion about the agent state.

\subsubsection*{Examples}

\textbf{1 = not similar}

\textit{True Response:} Hello, the annual sale began on January 23rd and ended on January 31st.

\textit{Generated Response:} Okay, let me look into that for you.

\textbf{3 = similar}

\textit{True Response:} I would be happy to find the answer for you.

\textit{Generated Response:} I would be happy to look into that for you.

\subsection{Relevance to Context?}
On a scale of 1 to 3 (1=not relevant, 3=relevant), how relevant is the generated response given the conversation context?

\subsubsection*{Scale}
\textbf{1 = not relevant}

Has nothing or very little to do with the conversation context.

\textbf{2 = somewhat relevant}

Is an OK response to the conversation context, though maybe missing some details or superfluous in some respects.

\textbf{3 = relevant}

A good response. If a customer saw this, they could believe a human wrote this. Note that this can include specific utterances like ‘OK, I booked that flight for you’ to generic ‘you’re welcome’, ‘please wait one moment’ if they make sense in that context.

\subsubsection*{Examples}
\textbf{1 = not relevant}

\textit{Conversation Context:} [REP] Hi, how may I help you this morning? 
[CUS] Yea, I had a quick question. I was checking my email and it says my subscription was removed. Is that true?  I still want it there.
[REP] Sure, I can check that for you. What is your account ID?
[CUS] Umm, not sure.

\textit{Generated Response:} What is the shipping status of the order?

\textbf{3 = relevant}

\textit{Conversation Context:} [REP] Hi, how may I help you this morning? 
[CUS] Yea, I had a quick question. I was checking my email and it says my subscription was removed. Is that true?  I still want it there.
[REP] Sure, I can check that for you. What is your account ID?
[CUS] Umm, not sure

\textit{Generated Response:} OK, and to whom do I have the pleasure of speaking with?

\section{Experimental Details}
\label{appx:experiment_details}

\textbf{Training Details}

\textbf{Stage 1. Train base TF model} We train / finetune a distilGPT2 / GPT2 medium model on all conversations for each dataset ABCD, MultiWoz, TaskMaster (separate model for each dataset) for 10 epochs. We call this the base teacher forcing (TF) model. We conducted grid search for the learning rate in \{1e-3, 5e-4, 2e-4, 1e-4, 5e-5, 2e-5, 1e-5\} and number of epochs in \{10, 20, 40\}.

\textbf{Stage 2. Generate Offline RL data} Given a TF model, we call it to generate an offline RL dataset. Each data point in the dataset is a tuple (context, response, reward). We obtain the reward by evaluating generated response against true response using thresholded BERTScore that we call BERTClick. For each context in the dataset, we include 1 true response + 5 model generated response.

\textbf{Stage 3. Train offline RL model} Finally, we fine tune the base TF model on the offline RL dataset to get models for each of the three offline RL methods TF Top, DT, ILQL. We implement TF, TF Top using the huggingface transformers library \cite{wolf2019huggingface} and ILQL using the trlx library \footnote{\url{https://github.com/CarperAI/trlx}}. We fine tune for 5 epochs. We disable gradients on context tokens during this stage of training so as to better match the inference time setup. Moreover, the same context repeats multiple times that would bias the gradients.

Training is done on an AWS EC2 g5.12xlarge instance which has 4 Nvidia A10G GPUs. For both Stage 1, 3 we pick the best epoch checkpoint based on the validation loss. We use the same set of hyper parameters across all datasets. More hyperparameter details in Tables \ref{tab:hyperparams_tf_dt}, \ref{tab:hyperparams_ilql} and \ref{tab:hyperparams_ppo}. On abcd dataset with 10k conversations using distilgpt2 model, training base model takes $\approx$20 mins for 10 epochs. Training offline RL model with 70\% subsamples takes, TF TOP: $\approx$2.5 hours for 5 epochs, DT: $\approx$6 hours for 5 epochs.

\begin{table}[!t]
\centering
\resizebox{0.8\columnwidth}{!}{
\begin{tabular}{llll}
\toprule
\textbf{Hyperparameters} & \textbf{TF} & \textbf{TF Top} & \textbf{DT} \\
\midrule
Model & DistilGPT / GPT2 Medium & DistilGPT / GPT2 Medium & DistilGPT / GPT2 Medium \\
Batch size & 16 / 32 & 32 / 64 & 32 / 64 \\
Block size & 1024 / 1024 & 512 / 512 & 512 / 512 \\
Max number of epochs & 10 & 5 & 5 \\
Optimizer & Adam & Adam & Adam \\
Learning rate &  1e-4 & 5e-5 & 5e-5 \\
Adam $(\beta_1, \beta_2)$ & (0.9, 0.999) & (0.9, 0.999)& (0.9, 0.999) \\
Adam $\epsilon$ & 1e-8 & 1e-8 & 1e-8 \\
Learning rate scheduler & Cosine decay & Cosine decay & Cosine decay \\
\bottomrule
\end{tabular}
}
\caption{TF, TF Top, DT training hyperparameters for ABCD, MultiWoz 2.2, TaskMaster-3 datasets. We tune the learning rate in \{5e-4, 1e-4, 5e-5, 1e-5\}.}
\label{tab:hyperparams_tf_dt}
\end{table}

\begin{table}[!t]
\centering
\begin{tabular}{ll}
\toprule
\textbf{Hyperparameters} & \textbf{ILQL} \\
\midrule
Model & DistilGPT\\
Batch size & 16 \\
Block size & 128 \\
Max number of iterations & 50000 \\
Optimizer & Adam \\
Learning rate &  1e-4 \\
Adam $(\beta_1, \beta_2)$ & (0.9, 0.95)\\
Adam $\epsilon$ & 1e-8 \\
Learning rate scheduler & Cosine \\
CQL Scale & 0.05 \\
$\tau$ & 0.7 \\
$\gamma$ & 0.99 \\
\bottomrule
\end{tabular}
\caption{ILQL Training hyperparameters for ABCD, MultiWoz 2.2, TaskMaster-3 datasets. We tune the learning rate in \{5e-4, 1e-4, 1e-5\} and CQL scale (regularization against base TF logits) in \{0.001, 0.005, 0.05, 0.5, 5\}}
\label{tab:hyperparams_ilql}
\end{table}

\begin{table}[!t]
\centering
\begin{tabular}{ll}
\toprule
\textbf{Hyperparameters} & \textbf{PPO} \\
\midrule
Model & DistilGPT\\
Batch size & 16 \\
Block size & 128 \\
Max number of iterations & 50000 \\
Optimizer & AdamW \\
Learning rate &  5e-7 \\
Adam $(\beta_1, \beta_2)$ & (0.9, 0.95)\\
Adam $\epsilon$ & 1e-8 \\
Weight decay & 1e-6 \\
Learning rate scheduler & Constant \\
PPO value coefficient & 2.3 \\
PPO KL initial coefficient & 0.2 \\
\bottomrule
\end{tabular}
\caption{PPO Training hyperparameters for ABCD, MultiWoz 2.2, TaskMaster 3 datasets. We tune the learning rate in \{1e-4, 5e-5, 2e-5, 1e-5, 5e-6, 2e-6, 1e-6, 5e-7, 2e-7, 1e-7\}, PPO value coefficient in \{1, 2, 2.3, 3\}, and PPO KL initial coefficient in \{0.2, 1, 2\}.}
\label{tab:hyperparams_ppo}
\end{table}

\section{Qualitative Results}
\label{appx:qual_results}

We show qualitative predictions for the different methods (TF, TF Top, DT) across the ABCD, MultiWoz 2.2 and Taskmaster-3 datasets. It is worth noting that while the responses generated by TF may appear relevant to the context, they often do not match the true human utterance and typically repeat information already in context. Offline RL approaches address these limitations in the TF responses. This is consistent with what we observed in the human evaluation study where the difference between \textit{similarity} annotations was greater than the difference in \textit{relevance} annotations across the three methods.

\begin{figure*}[!t]
\centering
\includegraphics[width=\textwidth, trim = 0cm 5.8cm 0cm 0cm]{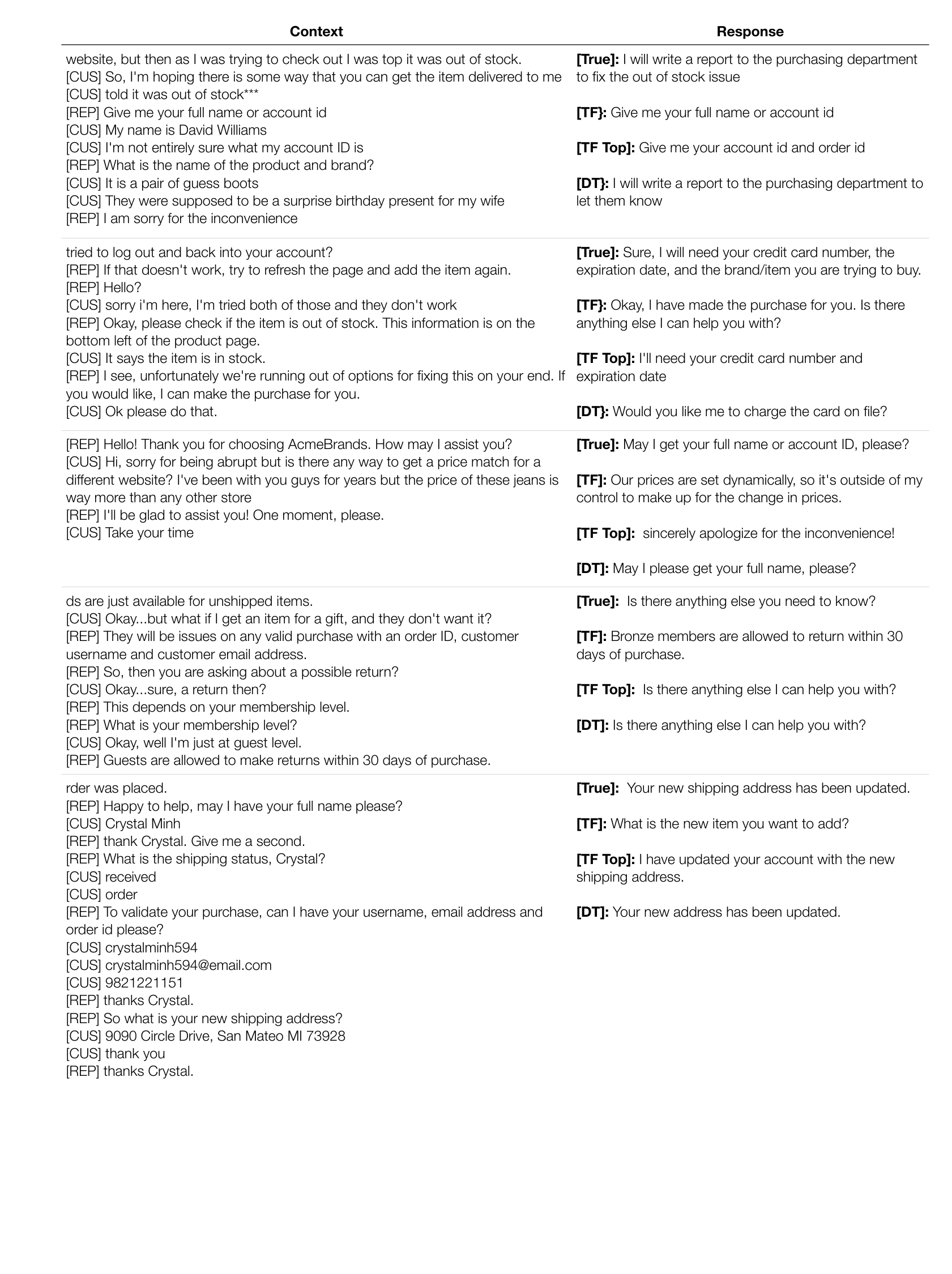}
\caption{{ABCD} Qualitative Predictions}
\vspace{-1em}
\label{fig:qual_examples_abcd}
\end{figure*}

\begin{figure*}[!t]
\centering
\includegraphics[width=\textwidth, trim = 0cm 5.0cm 0cm 0cm]{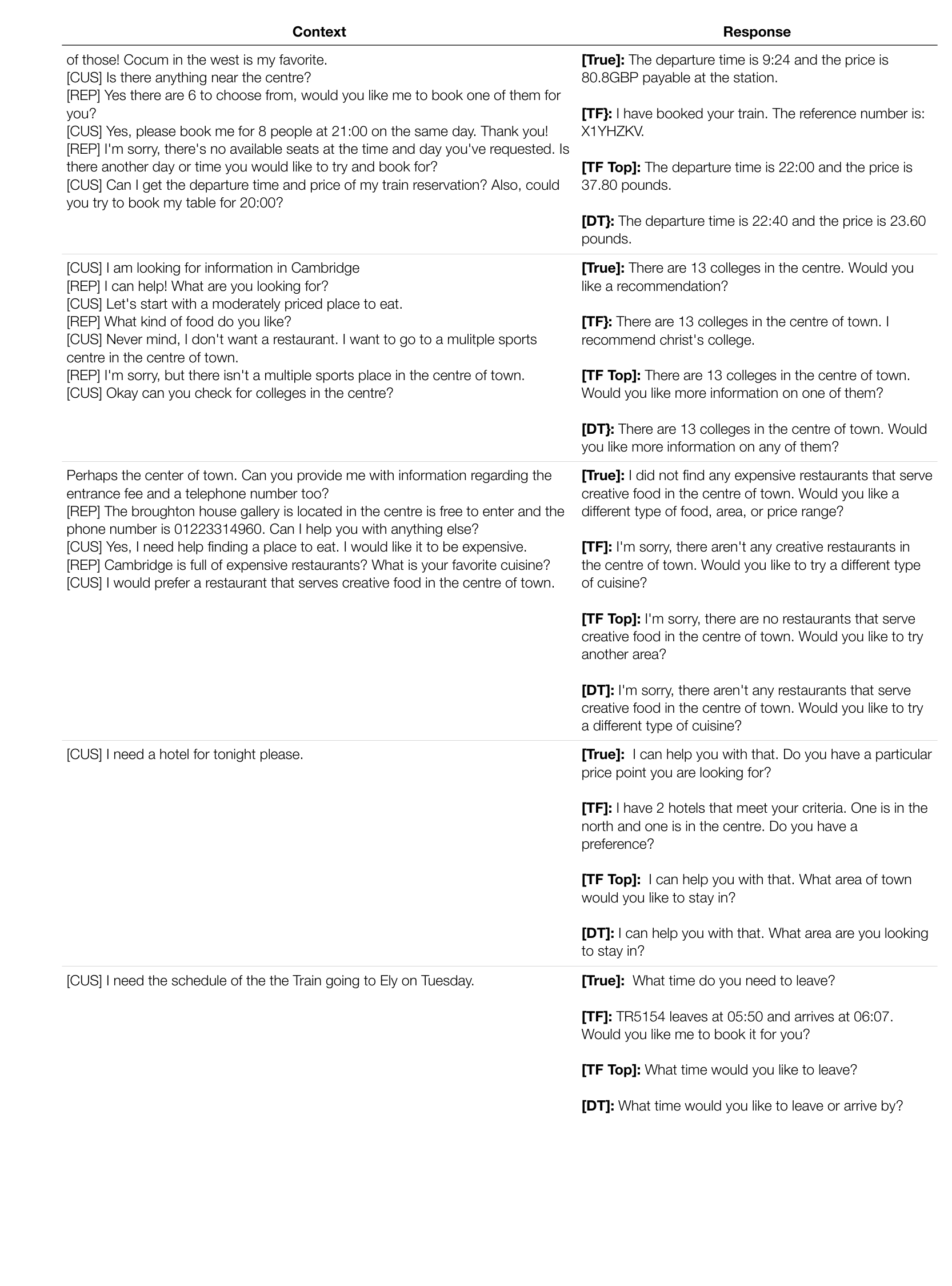}
\caption{MultiWoz 2.2 Qualitative Predictions}
\vspace{-1em}
\label{fig:qual_examples_multiwoz}
\end{figure*}

\begin{figure*}[!t]
\centering
\includegraphics[width=0.9\textwidth, trim = 0cm 4.5cm 0cm 0cm]{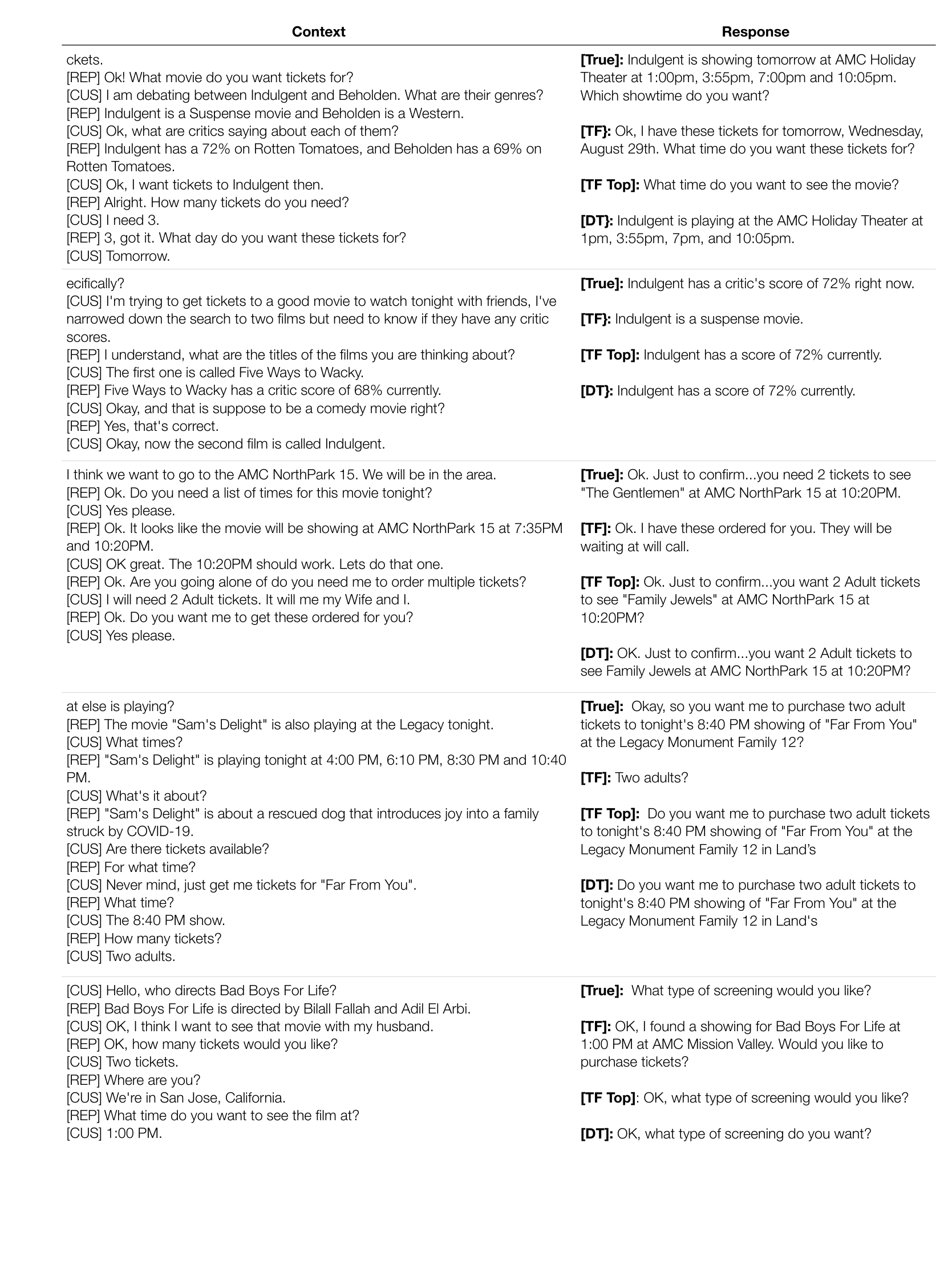}
\caption{Taskmaster-3 Qualitative Predictions}
\vspace{-1em}
\label{fig:qual_examples_taskmaster}
\end{figure*}

\clearpage
\section{Qualitative Results on Dialogue Metrics}

We also explored how \textbf{TF} and \textbf{TF-Top} perform against ground truth on dialogue metrics. The goal was to see whether these methods generate text that generates correct slot information. To do so, we trained a T5 dialogue state tracking (DST) model on MultiWoz. We then took generated responses and fed them to the DST model to compute slots and matched them with ground truth slots. Overall, we find that both \textbf{TF Top} and \textbf{TF} do worse than the ground truth, as expected. Ground truth utterances have access to privileged information which in turn defines the ground truth slots. For instance, a specific restaurant name that neither of the generated utterances would be able to predict ahead of time.

We note that we are not directly optimizing dialgoue metrics, so there is no guarantee that these metrics would improve with offline RL. However, we do observe that performance is not negatively affected. In fact, we see that \textbf{TF Top} scores higher than \textbf{TF} on the slot metrics. One such example is shown below. The context here is the user is asking the system to book a reservation in Zizzi Cambridge. The hallucinated hotel name in the TF response results in an incorrect slot value of "hotel-name": [ "alexander bed and breakfast" ]. 

\begin{figure*}[!htbp]
\centering
\includegraphics[width=0.9\textwidth]{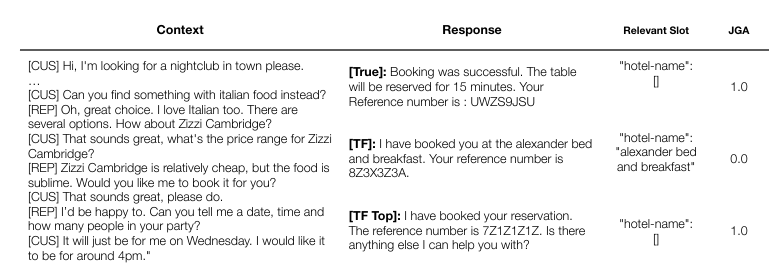}
\caption{Qualitative Predictions on MultiWoz to evaluate Dialogue Metrics}
\vspace{-1em}
\label{fig:qual_examples_dialogue}
\end{figure*}

\section{Additional Quantitative Results}

We also provide distribution over \textsc{BERTScores} for methods trained on 20\% of the data below. We find similar trends when models are trained on $80\%$ of the data, with slightly lower performance gains.

\begin{figure*}[!hbtp]
\centering
\includegraphics[width=\textwidth, trim = 0cm 0.8cm 0cm 0cm]{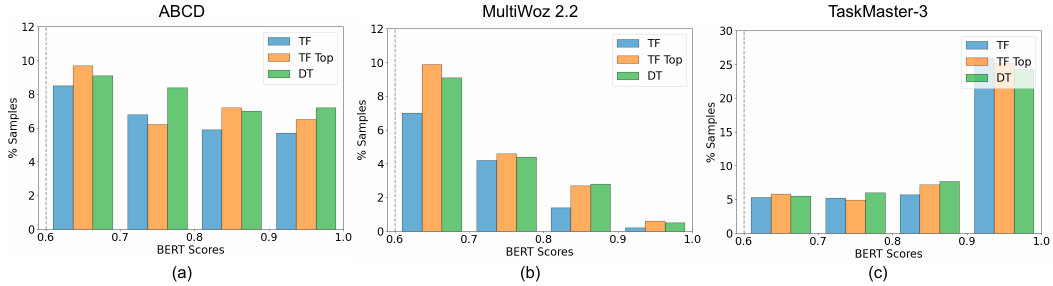}
\caption{Distribution over \textsc{BERTScores} for \textbf{(a)} ABCD \textbf{(b)} MultiWoz \textbf{(c)} Taskmaster-3 datasets with distilGPT2 finetuned on $20\%$ data.}
\vspace{-1em}
\label{fig:bert_score_hist}
\end{figure*}


\end{document}